\documentclass[10pt,twocolumn,letterpaper]{article}

\usepackage{algorithm}
\usepackage{algorithmic}
\usepackage{bm}
\usepackage[table]{xcolor}
\usepackage{graphicx}
\usepackage{makecell}
\usepackage{amsmath}
\usepackage{float}
\usepackage{hhline}
\usepackage{geometry}
\usepackage{booktabs}
\usepackage{arydshln}
\usepackage{amsfonts}
\usepackage{multirow}
\usepackage{subcaption}
\usepackage[table]{xcolor}  

\usepackage{pifont}
\newcommand{\cmark}{\ding{51}}  
\newcommand{\xmark}{\ding{55}}  
\usepackage{cvpr}              

\usepackage{tikz}
\usetikzlibrary{arrows.meta,positioning,calc,fit,shapes,backgrounds}

\definecolor{deepred}{RGB}{200,0,0}
\definecolor{deeporange}{RGB}{230,120,20}
\definecolor{deepgold}{RGB}{184,134,11}
\definecolor{deepgreen}{RGB}{0,128,0}
\definecolor{deepblue}{RGB}{0,0,180}
\definecolor{deepviolet}{RGB}{138,43,226}

\definecolor{cvprblue}{rgb}{0.21,0.49,0.74}
\usepackage[pagebackref,breaklinks,colorlinks,
            citecolor=cvprblue,
            linkcolor=cvprblue,
            urlcolor=cvprblue]{hyperref}
\renewcommand*{\backref}[1]{}
\renewcommand*{\backrefalt}[4]{%
  \ifcase #1 \or {\color{cvprblue}#2}\else {\color{cvprblue}#2}\fi}
\def\paperID{} 
\def\confName{CVPR}
\def\confYear{2026}

\title{ReflexSplit: Single Image Reflection Separation via Layer Fusion-Separation}

\author{Chia-Ming Lee$^{1,2}$ \quad Yu-Fan Lin$^{2}$\quad Jin-Hui Jiang$^{1}$\quad Yu-Jou Hsiao$^{2}$\\ \quad Chih-Chung Hsu$^{1,2}$ \quad Yu-Lun Liu$^1$\\
{$^1$National Yang Ming Chiao Tung University} \quad {$^2$National Cheng Kung University}}

\begin{document}


\maketitle

\begin{abstract}
Single Image Reflection Separation (SIRS) disentangles mixed images into transmission and reflection layers. 
Existing methods suffer from transmission-reflection confusion under nonlinear mixing, particularly in deep decoder layers, due to implicit fusion mechanisms and inadequate multi-scale coordination.
We propose ReflexSplit, a dual-stream framework with three key innovations.
(1) Cross-scale Gated Fusion (CrGF) adaptively aggregates semantic priors, texture details, and decoder context across hierarchical depths, stabilizing gradient flow and maintaining feature consistency.
(2) Layer Fusion-Separation Blocks (LFSB) alternate between fusion for shared structure extraction and differential separation for layer-specific disentanglement. Inspired by Differential Transformer, we extend attention cancellation to dual-stream separation via cross-stream subtraction.
(3) Curriculum training progressively strengthens differential separation through depth-dependent initialization and epoch-wise warmup.
Extensive experiments on synthetic and real-world benchmarks demonstrate state-of-the-art performance with superior perceptual quality and robust generalization. 
Our code is available at  \hyperlink{https://github.com/wuw2135/ReflexSplit}{https://github.com/wuw2135/ReflexSplit}.

\end{abstract}
\vspace{-0.5cm}
\section{Introduction}
\label{sec:intro}
 
\begin{figure}[t]
    \centering
\includegraphics[width=0.48\textwidth]{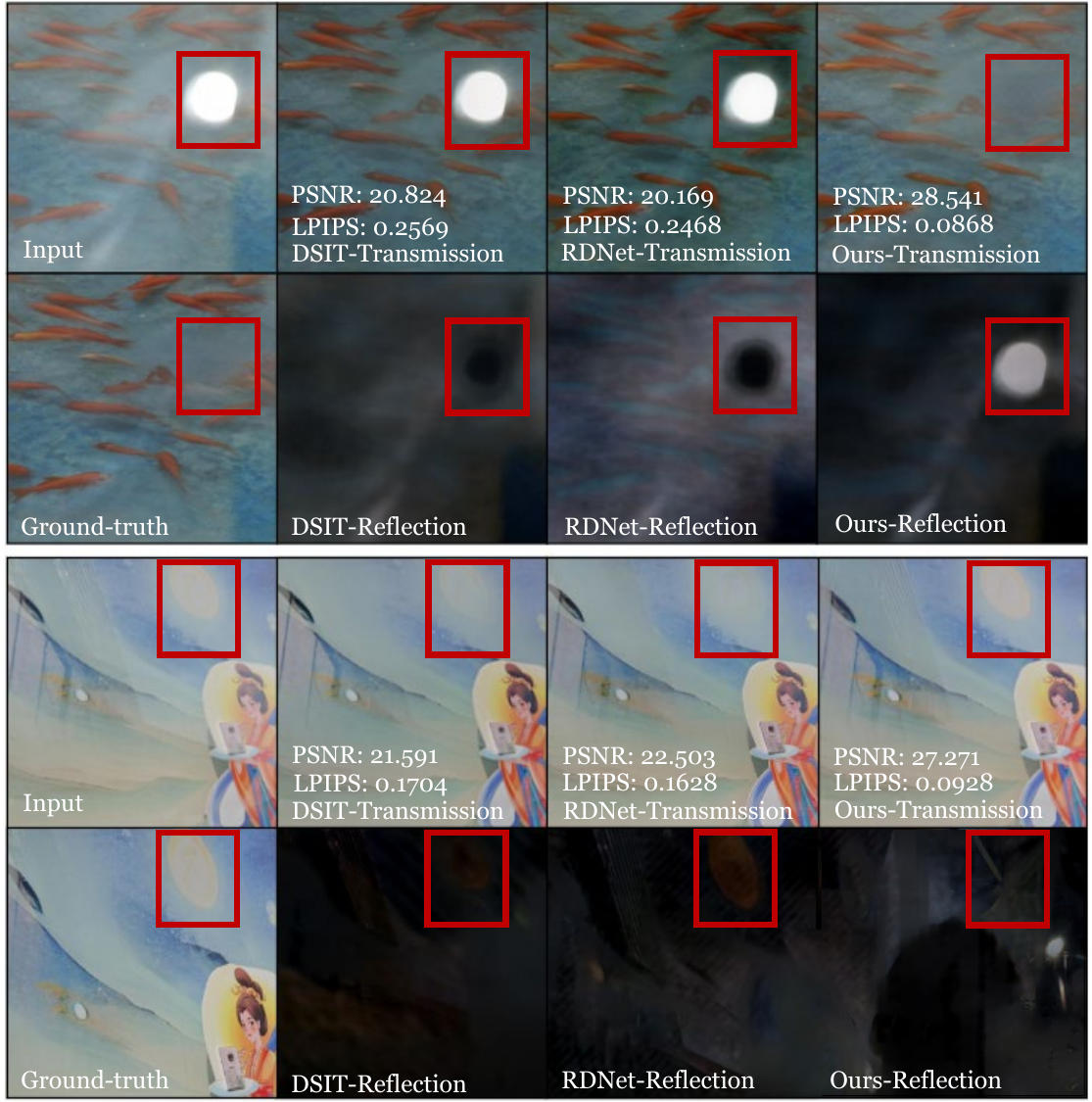}
    \caption{
    \textbf{Challenging examples on OpenRR-1K~\cite{openrr1k}.} 
    Both DSIT~\cite{hu2024single} and RDNet~\cite{zhao2024reversible} exhibit \textit{transmission-reflection confusion} with incomplete reflection separation, leading to annoying artifacts or details distortion. 
    ReflexSplit achieves better separation through explicit fusion-separation and multi-scale coordination.
    }
\label{fig:teaser}
    \vspace{-0.6cm}
\end{figure}

Reflection artifacts from transparent media degrade image quality, affecting applications like autonomous driving \cite{wang2024dc} and industrial inspection \cite{10637317}. Single Image Reflection Separation (SIRS) disentangles $\mathbf{I}$ into transmission $\mathbf{T}$ and reflection $\mathbf{R}$ layers. Unlike reflection removal treating reflection as noise, SIRS must recover both layers with distinct characteristics, requiring effective disentanglement to prevent inter-layer confusion \cite{zhang2018single}. Early work modeled this as $\mathbf{I} = \mathbf{T} + \mathbf{R}$~\cite{fan2017generic}, but this ignores layer asymmetry and complementarity, limiting real-world handling.

Subsequent work introduced weighted linear models $\mathbf{I} = \alpha \mathbf{T} + \beta \mathbf{R}$~\cite{wan2018crrn}, partially capturing layer asymmetry but relying on fixed global weights. More recently, Hu~\etal~\cite{hu2023single} proposed the nonlinear residual formulation
\begin{equation}
\mathbf{I} = \mathbf{T} + \mathbf{R} + \Phi(\mathbf{T}, \mathbf{R}),
\end{equation}
where $\Phi(\cdot)$ is a learnable nonlinear function modeling intricate inter-layer interactions. To better handle these nonlinear effects and achieve more effective separation of transmission and reflection layers, various methods have been proposed. YTMT~\cite{hu2021trash} and DSRNet~\cite{hu2023single} enhance inter-layer interaction through activation functions and channel splitting. DSIT~\cite{hu2024single} strengthens intra-layer feature interaction via dual-stream attention mechanisms. RDNet~\cite{zhao2024reversible}, inspired by the hypercolumn network \cite{hypercolumn}, designs a reversible decoupling network \cite{NIPS2017_f9be311e} to integrate and propagate multi-scale features, ensuring stable gradient propagation within the entire network—though at the cost of computational complexity and a two-stage training difficulty.

\begin{figure}[t]
    \centering
    \includegraphics[width=0.45\textwidth]{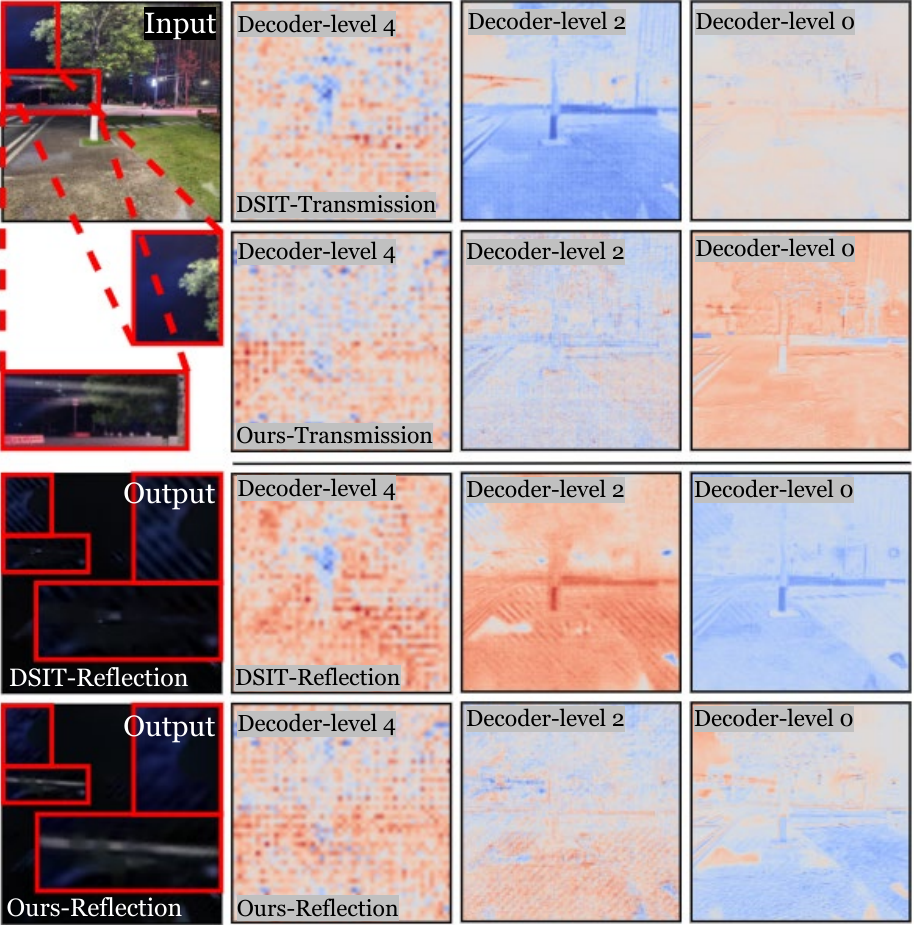}
\caption{
    \textbf{Layer-wise feature disentanglement comparison.}  DSIT~\cite{hu2024single} suffers from progressive transmission-reflection confusion in deep layers, exhibiting blurred boundaries and residual transmission leakage. Our method maintains clear layer distinction across all depths, preventing feature entanglement and achieving consistent reflection suppression.
}
    \label{fig:layer_confusion}
    \vspace{-0.6cm}
\end{figure}

\textbf{Despite these advances, we observe a persistent phenomenon:} when encountering strong reflections (e.g., intense light source reflections in water pools as shown in Figure~\ref{fig:teaser}(a)) or ambiguous scenarios (e.g., a moon painting on a wall being misidentified as reflection in Figure~\ref{fig:teaser}(b)), networks incorrectly confuse transmission and reflection layers (\emph{transmission-reflection confusion}), leading to suboptimal performance. Moreover, as commonly observed, increasing network depth often accompanies information loss, causing intra- and inter-layer features to become inseparable \cite{zhao2024reversible,drct}. This not only weakens the ability to model nonlinear effects but also exacerbates transmission-reflection confusion, particularly in deep decoder layers (Figure~\ref{fig:layer_confusion}).

To address these fundamental challenges, we propose \textbf{ReflexSplit}, a dual-stream framework with three key innovations:(1) Cross-scale Gated Fusion (CrGF): Unlike MuGI~\cite{hu2023single} which exchanges features at individual scales, CrGF adaptively aggregates semantic priors (GFEB), texture details (LFEB), and decoder context across hierarchical depths, stabilizing gradient flow and preventing feature degradation that causes transmission-reflection confusion. (2) Layer Fusion-Separation Block (LFSB): LFSB alternates between fusion and separation. \emph{Early fusion} shares complementary information via bidirectional projection, while \emph{Differential Dual-Dimensional Attention} performs cross-stream subtraction $\mathbf{A}^t - \lambda\mathbf{A}^r$ to suppress interference across spatial and inter-layer dimensions, ensuring layer-specific disentanglement.
(3) Curriculum Training: Differential separation is progressively strengthened through depth-dependent initialization and epoch-wise warmup, enabling the network to learn holistic reconstruction before focusing on layer-specific separation. 
As shown in Figure~\ref{fig:teaser}, ReflexSplit achieves superior separation under challenging nonlinear mixing where competing methods suffer from severe confusion. Our contributions are:
\begin{itemize}[leftmargin=*,noitemsep]
    \item \textbf{Explicit layer fusion-separation framework:} A paradigm that alternates between fusion for shared degradation extraction and differential separation for layer-specific disentanglement, preventing transmission-reflection confusion throughout the network hierarchy.
    
    \item \textbf{Cross-scale gated fusion and differential attention:} CrGF adaptively aggregates multi-scale features to maintain hierarchical consistency, while LFSB performs cross-stream attention subtraction for explicit layer disentanglement, stabilized by curriculum training.
    
    \item \textbf{State-of-the-art performance:} Superior results on both synthetic and real-world benchmarks, demonstrating robust layer separation and strong generalization capability.
\end{itemize}

\section{Related Work}
\label{sec:relatedworks}

\textbf{Physical Models for Reflection Superposition.}
Early reflection removal studies~\cite{fan2017generic, li2014single, levin2007user} 
relied on the linear superposition model $\mathbf{I} = \mathbf{T} + \mathbf{R}$, 
which assumes simple additive mixing between transmission and reflection layers.
To capture asymmetric contributions, 
weighted models such as $\mathbf{I} = \alpha \mathbf{T} + \beta \mathbf{R}$~\cite{wan2018crrn} 
and $\mathbf{I} = \alpha \mathbf{T} + (1-\alpha)\mathbf{R}$~\cite{yang2018seeing} 
introduced global attenuation coefficients $\alpha, \beta$.
However, these global weights fail to model spatial non-uniformity.

To address this limitation, spatially-varying models emerged.
Dong et al.~\cite{dong2021location} and Wen et al.~\cite{wen2019single} 
employed pixel-wise mixing weights $\mathbf{W}$, 
e.g., $\mathbf{I} = \mathbf{W} \circ \mathbf{T} + (1-\mathbf{W}) \circ \mathbf{R}$, 
enabling adaptive blending across spatial locations.
Beyond linear frameworks, 
Zheng et al.~\cite{zheng2021single} modeled absorption and refraction via spatially-varying coefficients,
while Wan et al.~\cite{wan2020reflection} introduced nonlinear mappings 
$\mathbf{I} = g(\mathbf{T}_s) + f(\mathbf{R}_s)$ to learn distinct degradation processes per layer.
Most recently, DSRNet~\cite{hu2023single} proposed the residual formulation 
$\mathbf{I} = \tilde{\mathbf{T}} + \tilde{\mathbf{R}} + \varphi(\mathbf{T}, \mathbf{R})$,
where a learnable residual term $\varphi(\mathbf{T},\mathbf{R})$ captures complex nonlinear inter-layer interactions.
\begin{figure*}[t]
    \centering
    \includegraphics[width=0.90\textwidth]{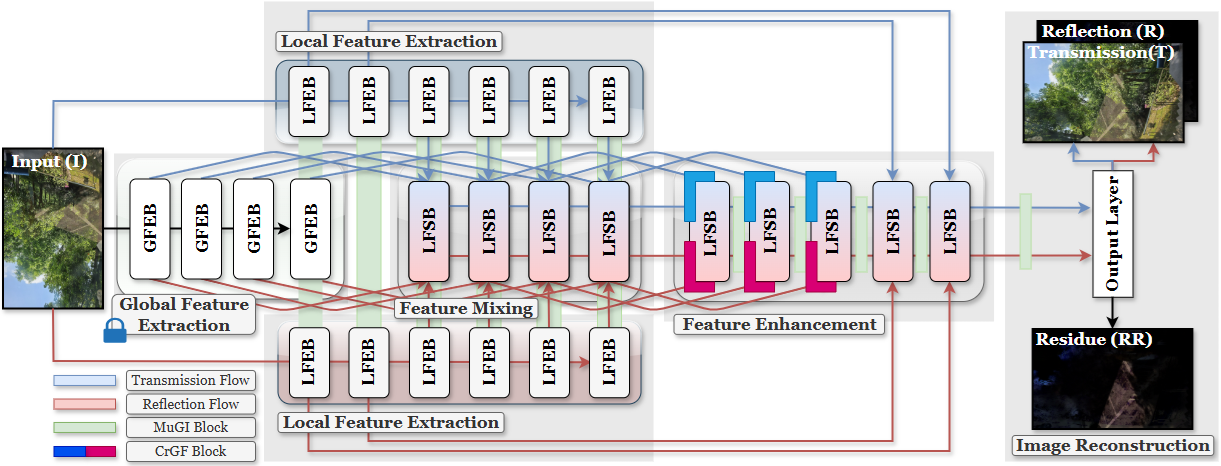}
    \caption{
    \textbf{Overview of ReflexSplit.} 
    Dual-branch encoders extract hierarchical features: GFEB (Swin Transformer) for global semantics $\{\mathbf{P}_\ell\}$ and LFEB (MuGI-based) for local textures $\{\mathbf{E}_\ell\}$. CrGF (Section~\ref{sec:crgf}) adaptively aggregates these multi-scale features across decoder depths, while LFSB (Section~\ref{sec:lfsb}) alternates between fusion (cross-stream complementarity) and differential separation (layer disentanglement) to progressively refine dual streams. Curriculum training (Section~\ref{sec:training}) progressively strengthens separation via depth-dependent initialization and epoch-wise warmup. Outputs: $\mathbf{T}$, $\mathbf{R}$, and residual $\mathbf{RR}$, which captures nonlinear interactions.
    }
    \label{fig:overview}
    \vspace{-0.4cm}
\end{figure*}
\vspace{-2mm}

\paragraph{Reflection Separation.}
Early multi-image methods exploited auxiliary cues—polarization~\cite{nayar1997separation, kong2011high, lei2020polarized}, focus~\cite{farid1999separating}, stereo~\cite{sinha2012image}, flash~\cite{agrawal2005removing, lei2021robust}, or motion~\cite{xue2015computational,liu2020learning}—to reduce ill-posedness, but required specialized equipment or controlled capture, limiting practical deployment. This motivated \emph{Single Image Reflection Separation (SIRS)}, which must disentangle layers from a single observation without auxiliary cues, making explicit inter-layer modeling critical.

Early SIRS works~\cite{li2023two, wan2019corrn, feng2021deep} employed multi-stage refinement with implicit complementarity, leading to suboptimal feature exchange. Recent methods formalized explicit interaction: YTMT~\cite{hu2021trash} exchanged suppressed features via ReLU operations; DSRNet~\cite{hu2023single} introduced Mutual Gated Interaction (MuGI) for dynamic cross-stream control; DSIT~\cite{hu2024single} combined self- and cross-attention but directly aggregates outputs without separation constraints, causing progressive confusion in deep layers (Figure~\ref{fig:layer_confusion}). RDNet~\cite{zhao2024reversible} uses reversible encoders for lossless flow but lacks explicit scale balancing, while DExNet~\cite{huang2025lightweight} achieves efficiency through sparse updates at the cost of expressiveness. RobustSIRR~\cite{robustsirr} enhances robustness via multi-scale attention and feature fusion with adversarial training, but it does not address hierarchical feature entanglement. Complementary approaches explored physical priors—HGRR~\cite{zhu2023hue} (hue maps), IBCLN~\cite{li2020single} (residual maps), Lei et al.~\cite{lei2020polarized} (polarization), PromptRR~\cite{zhong2024language} (multimodal guidance)—yet assume linear mixing and uniform attenuation, failing under over-exposure and specular highlights. DAI~\cite{dai} deliverer strong results using diffusion priors but at a prohibitive computational cost.

Despite progress, existing methods face two key challenges: inadequate hierarchical feature aggregation causing gradient instability, and implicit fusion mechanisms leading to progressive layer confusion. Our ReflexSplit tackles these through CrGF—which adaptively aggregates multi-scale features—and LFSB—which explicitly enforces layer disentanglement via differential attention.
\section{Methodology}
\label{sec:methodology}
Given a mixed image $\mathbf{I} \in \mathbb{R}^{H \times W \times 3}$, ReflexSplit decomposes it into transmission $\mathbf{T}$ and reflection $\mathbf{R}$ layers through explicit layer fusion-separation. As illustrated in Figure~\ref{fig:overview}, our dual-stream architecture comprises three core components: \textbf{(1)} Dual-branch feature extraction combining global semantic priors and local texture details to capture complementary multi-scale representations; \textbf{(2)} Cross-scale Gated Fusion (CrGF) that adaptively aggregates hierarchical features across decoder depths while stabilizing gradient flow; \textbf{(3)} Layer Fusion-Separation Blocks (LFSB) that alternate between adaptive fusion for shared structure extraction and differential separation for layer-specific disentanglement, preventing transmission-reflection confusion throughout the decoding hierarchy.

\paragraph{Dual-branch Feature Extraction.}
Following DSIT~\cite{hu2024single}, we separate global semantic modeling from local detail preservation through dual-branch extraction. A pretrained Swin Transformer~\cite{liu2021swin} serves as the Global Feature Extractor Block (GFEB), extracting semantic priors $\{\mathbf{P}_2, \mathbf{P}_3, \mathbf{P}_4, \mathbf{P}_5\}$ at multiple scales. Complementarily, a MuGI-based~\cite{hu2023single} CNN serves as the Local Feature Extractor Block (LFEB), capturing texture details $\{\mathbf{E}_0, \mathbf{E}_1, \mathbf{E}_2, \mathbf{E}_3, \mathbf{E}_4, \mathbf{E}_5\}$ at hierarchical resolutions $H_\ell = H/2^{\ell}$, $W_\ell = W/2^{\ell}$. This dual-branch design mitigates feature entanglement inherent in single-stream encoders, establishing complementary representations for subsequent fusion and separation.
\subsection{Cross-scale Gated Fusion}
\label{sec:crgf}
Stable global feature flow is critical for preventing progressive feature degradation and transmission-reflection confusion. While intra-layer mechanisms like MuGI~\cite{hu2023single} enhance complementarity at individual scales, adaptive cross-scale coordination from heterogeneous sources (semantic priors, texture details, decoder context) is equally essential. Existing methods fall short: DSIT~\cite{hu2024single} suffers gradient instability; RobustSIRR~\cite{robustsirr} performs multi-scale fusion via direct concatenation without adaptive gating; RDNet~\cite{zhao2024reversible} lacks explicit scale coordination; MuGI operates only at single scales. Without adaptive cross-scale coordination, these methods struggle to balance semantic richness and fine-grained details across hierarchical depths, leading to progressive feature degradation.

\begin{figure}[t]
\centering
\includegraphics[width=\linewidth]{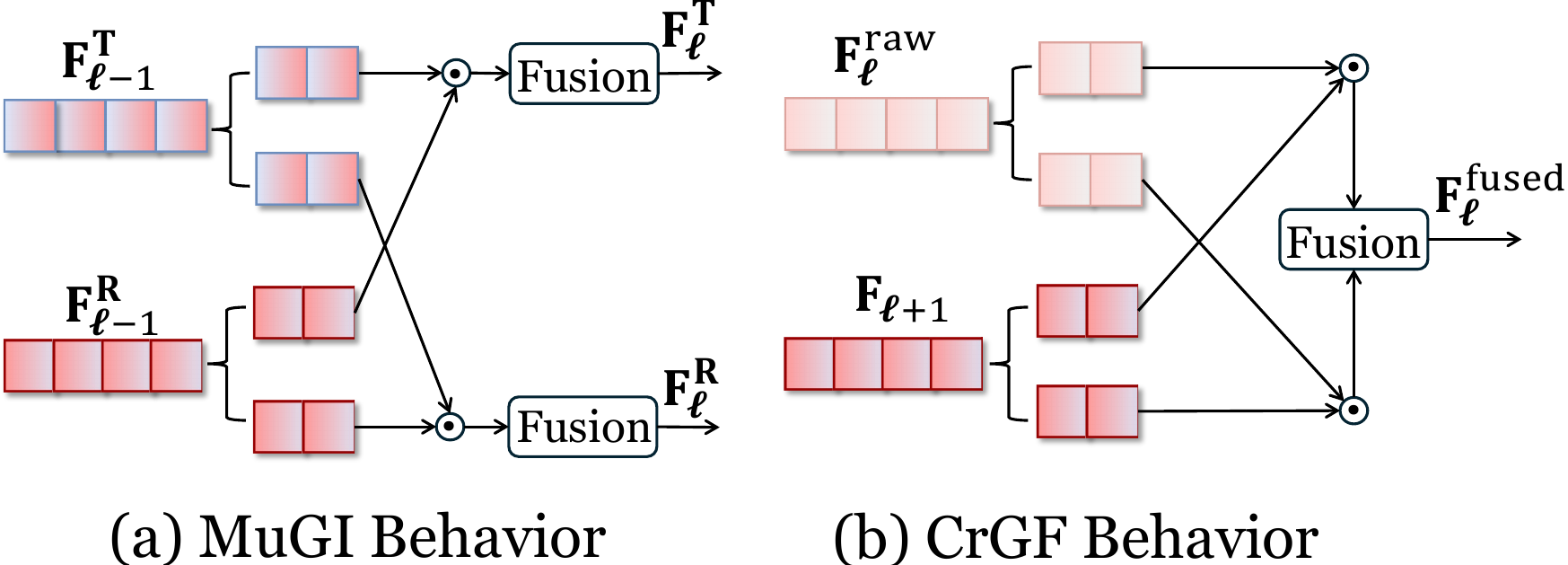}
\caption{\textbf{Complementary roles of MuGI and proposed CrGF.} 
(a) MuGI~\cite{hu2023single} focuses on dual-stream \emph{feature interaction}: transmission and reflection features exchange information via channel-wise gating at each decoder level.
(b) CrGF focuses on multi-scale \emph{feature integration}: adaptively aggregating hierarchical encoder features and decoder context across scales to maintain gradient stability and feature consistency throughout decoding.}
\label{fig:crgf_comparison}
\vspace{-4mm}
\end{figure}

To enable flexible recombination of multi-scale, multi-source features, we propose CrGF at decoder Levels $\{4, 3, 2\}$. As illustrated in Figure~\ref{fig:crgf_comparison}, CrGF extends MuGI's gating principle to cross-scale aggregation via bidirectional paths that adaptively balance decoder context ($\mathbf{F}_{\ell+1}$), semantic priors ($\mathbf{P}_\ell$), and texture details ($\mathbf{E}_\ell$):
\begin{equation}
\label{eq:gated_fusion}
\left\{
\begin{aligned}
\mathbf{F}_\ell^{\text{main}} &= \mathcal{G}_1(\mathbf{F}_\ell^{\text{raw}}) \odot \mathcal{G}_2(\mathbf{F}_{\ell+1}); \\
\mathbf{F}_\ell^{\text{aux}} &= \mathcal{G}_1(\mathbf{F}_{\ell+1}) \odot \mathcal{G}_2(\mathbf{F}_\ell^{\text{raw}}),
\end{aligned}
\right.
\end{equation}
where $\mathbf{F}_\ell^{\text{raw}} = \mathbf{F}_{\ell+1} + \mathbf{P}_\ell + \mathbf{E}_\ell$, and $\mathcal{G}_{1,2}$ select complementary channels via splitting. $\mathbf{F}_\ell^{\text{main}}$ emphasizes current-level features gated by context, while $\mathbf{F}_\ell^{\text{aux}}$ propagates context gated by current cues, enabling bidirectional information flow. Final fusion:
\begin{equation}
\mathbf{F}_\ell^{\text{fused}} = w_\ell^{(1)} \phi_1(\mathbf{F}_\ell^{\text{main}}) + w_\ell^{(2)} \phi_2(\mathbf{F}_\ell^{\text{aux}}),
\end{equation}
where $\phi_{1,2}$ are $1 \times 1$ convolutions and $w_\ell$ normalized by softmax. CrGF's adaptive bidirectional gating dynamically selects and recombines features based on context, addressing the limitations of RobustSIRR's static concatenation and RDNet's fixed reversible paths. This enables flexible cross-scale coordination that prevents progressive degradation and establishes well-structured representations for LFSB's layer disentanglement. At Levels $\{1, 0\}$, due to the absence of extracted global feature $\mathbf{P}_\ell$, we simply use $\mathbf{F}_\ell = \mathbf{F}_{\ell+1} + \mathbf{E}_\ell$ for feature aggregation.

\begin{figure}[t]
    \centering\includegraphics[width=\linewidth]{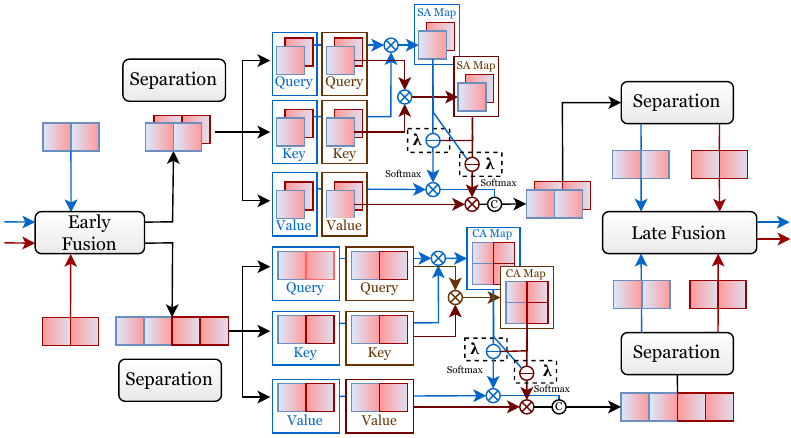}
    \caption{
        \textbf{Layer Fusion-Separation Block (LFSB).}
        LFSB alternates between fusion (shared structure) and separation (layer disentanglement): 
        (1) Bidirectional projection aligns transmission-reflection features;
        (2) Dual-dimensional attention (SA + CA) models spatial and inter-layer dependencies;
        (3) Differential operators $\mathbf{A}^{t} - \lambda_\ell \mathbf{A}^{r}$ suppress cross-stream interference;
        (4) FFNs with residuals integrate separated features.
    }
    \label{fig:lfsb}
    \vspace{-0.5cm}
\end{figure}

\subsection{Layer Fusion-Separation Block}
\label{sec:lfsb}
Complementary to CrGF's global feature flow, explicit \textit{intra-layer disentanglement} is critical for preventing transmission-reflection confusion. However, existing methods lack effective separation mechanisms: DSRNet~\cite{hu2023single} uses MuGI without attention-based separation; RDNet~\cite{zhao2024reversible} preserves flow via reversibility but lacks explicit complementarity modeling; DSIT~\cite{hu2024single} aggregates dual-dimensional attention without separation constraints, causing progressive feature confusion (Figure~\ref{fig:layer_confusion}).

To address these limitations, we propose LFSB (Figure~\ref{fig:lfsb}), which alternates between two complementary operations—fusion for extracting shared structure and differential dual-dimensional attention for enforcing layer-specific disentanglement.

\paragraph{Early Fusion.}
Before applying attention, we align semantic spaces via bidirectional cross-stream projection to establish a foundation for effective layer interaction.
At decoder stage $\ell$, given $\mathbf{F}^{t}_{\ell}, \mathbf{F}^{r}_{\ell} \in \mathbb{R}^{H_\ell \times W_\ell \times C}$:
\begin{equation}
\mathbf{F}^{t\prime}_{\ell} = \mathbf{W}^{t}[\mathbf{F}^{t}_{\ell} \,\|\, \mathbf{F}^{r}_{\ell}], \quad
\mathbf{F}^{r\prime}_{\ell} = \mathbf{W}^{r}[\mathbf{F}^{r}_{\ell} \,\|\, \mathbf{F}^{t}_{\ell}],
\end{equation}
where $\mathbf{W}^{t}, \mathbf{W}^{r} \in \mathbb{R}^{C \times 2C}$ extract shared degradation patterns between transmission and reflection layers,
enabling each stream to benefit from complementary cues before feature separation.
This bidirectional projection explicitly aligns feature spaces, preparing well-structured representations for subsequent differential attention computation.

\paragraph{Differential Dual-Dimensional Attention.}
Following DSIT~\cite{hu2024single}, we employ dual-dimensional attention to jointly model intra-layer spatial correlations and inter-layer complementarity after window-partitioning.
Inspired by Differential Transformer~\cite{diff} that suppresses noise through attention cancellation in single-stream architectures,
we extend this principle to dual-stream layer separation by introducing differential operators across transmission and reflection branches.
However, unlike Differential Transformer that computes $\text{Attn}_{\text{diff}} = \text{softmax}(\mathbf{Q}\mathbf{K}^\top / \sqrt{d}) - \lambda \cdot \text{softmax}(\mathbf{Q}\mathbf{K}^\top / \sqrt{d})$ within the same attention head to cancel noise patterns,
our method performs cross-stream subtraction $\mathbf{A}^{t} -  \lambda_\ell \mathbf{A}^{r}$ to suppress inter-layer interference between complementary transmission and reflection features.

We concatenate features along the batch dimension for self-attention (SA)
$\mathbf{F}_{\text{SA}} = \text{concat}([\mathbf{F}^{t\prime}_{\ell}, \mathbf{F}^{r\prime}_{\ell}]) \in \mathbb{R}^{2B \times N \times C}$,
computing independent attention maps $\mathbf{A}^{t}_{\text{SA}}, \mathbf{A}^{r}_{\text{SA}}$ for spatial refinement.
Complementarily, we concatenate along the sequence dimension for cross-attention (CA) 
$\mathbf{F}_{\text{CA}} = \text{concat}([\mathbf{F}^{t\prime}_{\ell}, \mathbf{F}^{r\prime}_{\ell}]) \in \mathbb{R}^{B \times 2N \times C}$,
capturing inter-layer dependencies via $\mathbf{A}^{t}_{\text{CA}}, \mathbf{A}^{r}_{\text{CA}}$.

\begin{figure}
    \centering
    \includegraphics[width=0.45\textwidth]{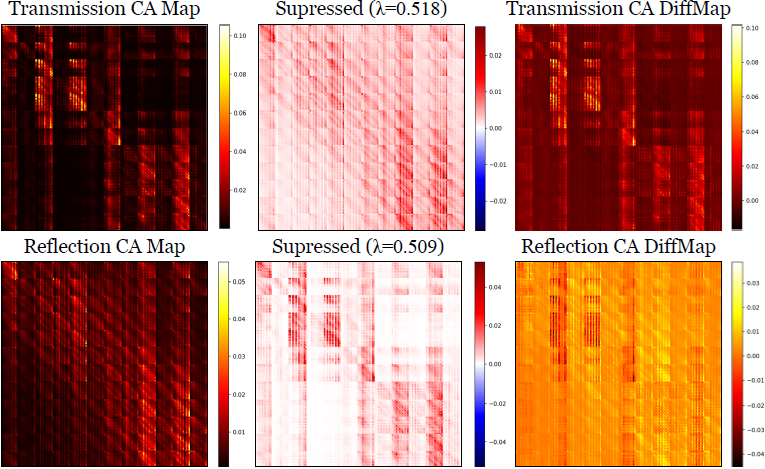}
\caption{\textbf{Differential attention visualization.}
    Cross-attention maps reveal overlapping patterns between transmission and reflection (left). 
    Differential separation via $\mathbf{A}^t - \sigma(\lambda_\ell)\mathbf{A}^r$ actively suppresses cross-stream interference (middle), 
    producing balanced, layer-specific attention distributions (right) that enable effective feature disentanglement.
}
    \label{fig:diff}
    \vspace{-0.4cm}
\end{figure}

Afterwards, unlike DSIT that directly aggregates SA and CA outputs, we introduce differential operators to actively suppress cross-stream interference:
\begin{equation}
\begin{aligned}
\mathbf{A}^{t}_{\text{diff}} &= (\mathbf{A}^{t}_{\text{SA}} + \mathbf{A}^{t}_{\text{CA}}) - \sigma(\lambda_{\ell}) (\mathbf{A}^{r}_{\text{SA}} + \mathbf{A}^{r}_{\text{CA}}); \\
\mathbf{A}^{r}_{\text{diff}} &= (\mathbf{A}^{r}_{\text{SA}} + \mathbf{A}^{r}_{\text{CA}}) - \sigma(\lambda_{\ell}) (\mathbf{A}^{t}_{\text{SA}} + \mathbf{A}^{t}_{\text{CA}}),
\end{aligned}
\end{equation}
where learnable $\lambda_{\ell}$ in $[0,1]$, controls differential strength and $\sigma(\cdot)$ is sigmoid function.
This cross-stream subtraction $-\sigma(\lambda_{\ell}) \mathbf{A}^{r}$ ensures transmission-specific and reflection-specific features remain distinguishable throughout decoder stages, 
addressing the progressive feature confusion.
As shown in Figure~\ref{fig:diff}, this differential mechanism effectively separates layer-specific attention patterns by distributing attention across multiple heads, reducing noisy activations and enabling clearer distinction between transmission and reflection features.

\paragraph{Late Fusion \& Aggregation.}
The separated attention features are integrated via feed-forward refinement with residual connections:
\begin{equation}
\mathbf{F}^{t}_{\ell+1} = \mathbf{F}^{t}_{\ell} + \text{FFN}(\mathbf{A}^{t}_{\text{diff}}), \quad
\mathbf{F}^{r}_{\ell+1} = \mathbf{F}^{r}_{\ell} + \text{FFN}(\mathbf{A}^{r}_{\text{diff}}),
\end{equation}
stabilizing gradient propagation across decoder depths.
Combined with our stage-wise curriculum training (Section~\ref{sec:training}), 
this explicit fusion-separation paradigm maintains clear transmission-reflection distinction throughout the network hierarchy.

\subsection{Curriculum Training and Loss Function}
\label{sec:training}
LFSB's differential separation relies on $\lambda_\ell$ to suppress cross-stream interference via $\mathbf{A}^{t} - \sigma(\lambda_\ell) \mathbf{A}^{r}$. However, $\lambda$ requires careful control: excessive strength in early training destabilizes poorly-structured features, while insufficient strength fails to enforce separation, causing confusion. We adopt curriculum learning~\cite{acl} to progressively strengthen differential separation, enabling the network to learn shared structure before refining layer-specific features, through two complementary mechanisms.

\noindent\textbf{Depth-Dependent Initialization.}  
Different decoder depths require different separation strengths.
For LFSB at decoder layer $\ell$:
\begin{equation}
\label{eq:lambda_init}
    \lambda_\ell^{\text{init}} = 0.8 - 0.6\, e^{-0.3 \ell},
\end{equation}
where deeper layers receive stronger weights ($\lambda \to 0.8$) for global separation,
while shallow layers maintain weaker weights ($\lambda \to 0.2$) to preserve fine-grained details.

\noindent\textbf{Epoch-Wise Warmup.}  
To stabilize training, we gradually increase the differential separation strength across epochs.
At epoch $e \in \{1, \ldots, E_{\text{total}}\}$, the global scaling factor is:
\begin{equation}
\label{eq:warmup_schedule}
\lambda_\text{diff}(e) = 
\begin{cases}
0.1 + 0.9 \dfrac{e}{E_\text{warmup}}, & e < E_\text{warmup},\\[0.5em]
1.0, & e \ge E_\text{warmup},
\end{cases}
\end{equation}
where $E_{\text{warmup}} = 30$ is the warmup period.
This schedule linearly increases $\lambda_\text{diff}(e)$ from 0.1 to 1.0 during the first 30 epochs,
then maintains full strength thereafter.
The final effective coefficient combines spatial (depth-dependent) and temporal (epoch-wise) adaptation:
\begin{equation}
\lambda_\ell(e) = \lambda_\ell^{\text{init}} \cdot \lambda_\text{diff}(e).
\end{equation}
Early training prioritizes holistic reconstruction with weak differential constraints,
while later stages enforce strong layer-specific separation.


\begin{table*}
    \centering
    \caption{\textbf{Quantitative comparisons on synthetic datasets.} 
    We evaluate on Real20, Nature, and three subsets of SIR$^2$~\cite{wan2017benchmarking} (Objects, Postcard, Wild). 
    Best, second-best, and third-best results are highlighted in \colorbox{red!25}{red}, \colorbox{orange!25}{orange}, and \colorbox{yellow!25}{yellow} respectively.}
    \label{tab:sirs_results}
    \resizebox{\textwidth}{!}{
    \begin{tabular}{llcc *{12}{c}}
        \toprule
        & \multirow{2}{*}{Methods} & \multirow{2}{*}{Venue}
        & \multirow{2}{*}{Trainable Parameters}
        & \multicolumn{2}{c}{Real20 (20)}
        & \multicolumn{2}{c}{Objects (200)}
        & \multicolumn{2}{c}{Postcard (199)}
        & \multicolumn{2}{c}{Wild (101)}
        & \multicolumn{2}{c}{Nature (20)}
        & \multicolumn{2}{c}{Average (540)} \\
        \cmidrule(lr){5-6}\cmidrule(lr){7-8}\cmidrule(lr){9-10}\cmidrule(lr){11-12}\cmidrule(lr){13-14}\cmidrule(lr){15-16}
        & & & &
        PSNR$\uparrow$ & SSIM$\uparrow$ &
        PSNR$\uparrow$ & SSIM$\uparrow$ &
        PSNR$\uparrow$ & SSIM$\uparrow$ &
        PSNR$\uparrow$ & SSIM$\uparrow$ &
        PSNR$\uparrow$ & SSIM$\uparrow$ &
        PSNR$\uparrow$ & SSIM$\uparrow$ \\
        \midrule
        & Zhang et al.~\cite{zhang2018single} & CVPR 2018 & 3.1M & 21.36 & 0.776 & 24.27 & 0.889 & 22.89 & 0.877 & 24.82 & 0.898 & 22.00	& 0.779	& 23.67	& 0.878 \\
        & ERRNet~\cite{wei2019single} & CVPR 2019 & 28.8M & 22.69 & 0.803 & 24.10 & 0.891 & 20.78 & 0.836 & 25.33 & 0.903 & 23.08 & 0.800 &	23.01 &	0.866 \\
        & IBCLN~\cite{li2020single} & CVPR 2020 & 24.3M & 19.42 & 0.750 & 24.01 & 0.884 & 22.77 & 0.871 & 23.52 & 0.878 & 20.68 & 0.763 & 23.17 &	0.869 \\
        & Zheng et al.~\cite{zheng2021single} & CVPR 2021 & 38.4M & 18.07 & 0.694 & 22.48 & 0.813 & 19.65 & 0.819 & 19.99 & 0.798 & 18.40 &	0.690 &	20.66 &	0.803 \\
        & YTMT~\cite{hu2021trash} & NeurIPS 2021 & 29.9M & 23.01 & 0.810 & 25.03 & 0.898 & 22.32 & 0.864 & 24.95 & 0.888 & 24.13 & 0.818 & 23.91 & 0.877 \\
        & DSRNet~\cite{hu2023single} & ICCV 2023 & 123.6M & 24.19 & 0.821 & 26.74 & 0.921 & 24.46 & 0.904 & 26.52 & 0.922 & 25.61 &	0.844 &	25.72 & 0.908 \\
        & Zhong et al.~\cite{zhong2024language} & CVPR 2024 & -- & 24.05 & 0.824 & 26.51 & \cellcolor{yellow!25}0.927 & \cellcolor{yellow!25}25.02 & \cellcolor{orange!25}0.915 & 26.23 & 0.925 & 23.87 & 0.812 & 25.72 & 0.914 \\
        & MaxRF~\cite{zhu2024revisiting} & CVPR 2024 & 27.9M & 21.58 & 0.794 & 25.80 & 0.915 & 21.45 & 0.863 & 26.35 & 0.924 & 25.58 &	0.837 & 24.14 & 0.890 \\
        & DSIT~\cite{hu2024single} & NeurIPS 2024 & 136M & \cellcolor{yellow!25}24.71 & \cellcolor{yellow!25}0.831 & \cellcolor{yellow!25}26.77 & 0.921 & 24.47 & \cellcolor{yellow!25}0.910 & \cellcolor{yellow!25}27.11 & \cellcolor{yellow!25}0.925 & \cellcolor{yellow!25}26.61 & \cellcolor{yellow!25}0.844 & \cellcolor{yellow!25}25.93 & \cellcolor{yellow!25}0.886 \\
        & DExNet~\cite{huang2025lightweight} & TPAMI 2025 & 9.6M & 23.42 & 0.816 & 25.37 & 0.909 & 24.21 & 0.906 & 26.65 & 0.922 & 23.39 &	0.828 &	25.04 &	0.904\\
        & RDNet~\cite{zhao2024reversible} & CVPR 2025 & 266.4M & \cellcolor{orange!25}25.17 & \cellcolor{orange!25}0.841 & \cellcolor{red!25}27.11 & \cellcolor{orange!25}0.925 & \cellcolor{orange!25}25.04 & \cellcolor{yellow!25}0.910 & \cellcolor{red!25}27.86 & \cellcolor{orange!25}0.931 & \cellcolor{orange!25}26.75 & \cellcolor{orange!25}0.846 & \cellcolor{orange!25}26.38 & \cellcolor{orange!25}0.890 \\
        \midrule
        & \textbf{ReflexSplit (Ours)} & -- & 174M & \cellcolor{red!25}\textbf{25.22} & \cellcolor{red!25}\textbf{0.846} & \cellcolor{orange!25}27.08 & \cellcolor{red!25}\textbf{0.929} & \cellcolor{red!25}\textbf{25.38} & \cellcolor{red!25}\textbf{0.927} & \cellcolor{orange!25}27.30 & \cellcolor{red!25}\textbf{0.933} & \cellcolor{red!25}\textbf{27.03} & \cellcolor{red!25}\textbf{0.854} & \cellcolor{red!25}\textbf{26.40} & \cellcolor{red!25}\textbf{0.898} \\
        \bottomrule
    \end{tabular}
    }
    \vspace{-0.3cm}
\end{table*}

\paragraph{Loss Functions.}
Let $\hat{\mathbf{T}}$, $\hat{\mathbf{R}}$, and $\hat{\mathbf{RR}}$ denote predicted transmission, reflection, and residual layers, 
with $\mathbf{T}$, $\mathbf{R}$ as ground truth. The training objective is:
\begin{equation}
\begin{aligned}
\mathcal{L}_{\text{total}} 
&= \lambda_\text{rec}\,\mathcal{L}_{\text{rec}} 
+ \lambda_\text{refl}\,\mathcal{L}_{\text{refl}} 
+ \lambda_\text{vgg}\,\mathcal{L}_{\text{vgg}}+
\lambda_\text{color}\,\mathcal{L}_{\text{color}}\\
&\quad + 
\lambda_\text{exclu}\,\mathcal{L}_{\text{exclu}}+ 
\lambda_\text{recons}\,\mathcal{L}_{\text{recons}},
\end{aligned}
\end{equation}
where we use Charbonnier loss~\cite{charbonier} $\mathcal{L}_{\text{rec}} = \sqrt{\|\hat{\mathbf{T}} - \mathbf{T}\|^2 + \epsilon^2}$ ($\epsilon = 10^{-6}$) 
and $\ell_1$ loss $\mathcal{L}_{\text{refl}} = \|\hat{\mathbf{R}} - \mathbf{R}\|_1$ for pixel-wise supervision.
$\mathcal{L}_{\text{vgg}}$ is VGG perceptual loss~\cite{vgg} using features from layers \{2, 7, 12, 21, 30\}. $\mathcal{L}_{\text{color}}$ is color consistency loss for output's reflection color enhancement.
$\mathcal{L}_{\text{exclu}}$ is exclusion loss~\cite{zhang2018single} encouraging gradient orthogonality between layers:
\begin{equation}
\mathcal{L}_{\text{exclu}} = \sum_{l=1}^{3} \left( \|\nabla_x \hat{\mathbf{T}} \odot \nabla_x \hat{\mathbf{R}}\|_1 + \|\nabla_y \hat{\mathbf{T}} \odot \nabla_y \hat{\mathbf{R}}\|_1 \right),
\end{equation}
where $\nabla_x$, $\nabla_y$ are spatial gradients and $\odot$ is element-wise product.
$\mathcal{L}_{\text{recons}} = \|\hat{\mathbf{T}} + \hat{\mathbf{R}} + \hat{\mathbf{RR}} - \mathbf{I}\|_1$ enforces reconstruction consistency.
Color consistency loss $\mathcal{L}_{\text{color}} = \|\mu(\hat{\mathbf{R}}) - \mu(\mathbf{R})\|_1 + \|\sigma(\hat{\mathbf{R}}) - \sigma(\mathbf{R})\|_1,$ is enforced to  maintain color fidelity in the
separated reflection layer. We set $\lambda_\text{rec} = 1.0$, $\lambda_\text{refl} = 0.5$, $\lambda_\text{vgg} = 0.1$, $\lambda_\text{exclu} = 1.0$, $\lambda_\text{recons} = 0.2$.
As $\lambda_\text{diff}(e)$ increases during training, 
the model shifts from holistic reconstruction to differential refinement.


\section{Experiment Results}
\label{sec:experimentSetup}
\noindent\textbf{Dataset and Implementation Details.} 
We train on a composite dataset comprising 
7,643 synthetic pairs from PASCAL VOC~\cite{everingham2010pascal}, 
90 real pairs from~\cite{zhang2018single}, 
and 200 pairs from Nature~\cite{li2020single}.
For synthetic data, transmission and reflection layers are randomly sampled from PASCAL VOC and blended following the nonlinear model inspired by the "screen" blending mode in digital image processing~\cite{hu2023single}:
\begin{equation}
    \mathbf{I}_{\text{syn}} = \gamma_1 \mathbf{T}_{\text{syn}} + \gamma_2 \mathbf{R}_{\text{syn}} - \gamma_1\gamma_2 \mathbf{T}_{\text{syn}} \circ \mathbf{R}_{\text{syn}},
\end{equation}
where $\mathbf{T}_{\text{syn}}$, $\mathbf{R}_{\text{syn}}$, and $\mathbf{I}_{\text{syn}}$ represent transmission, reflection, and superimposed layers during synthesis, respectively. 
The attenuation coefficients are randomly sampled as $\gamma_1 \in [0.8, 1.0]$ and $\gamma_2 \in [0.4, 1.0]$ to simulate varying layer intensities. 
This formulation reserves lighter colors for the blending layers, effectively modeling real-world nonlinear mixing phenomena such as over-exposure and specular highlights.
Each epoch samples 5,000 pairs with distribution ratio $0.6:0.2:0.2$ (synthetic:real:Nature).

For evaluation, we use five benchmarks: 
Real20 (20 images), Nature (20 images), 
and three SIR$^2$ subsets~\cite{wan2017benchmarking}— Objects (200), Postcard (199), and Wild (101).
We also evaluate on the real-world OpenRR-1K dataset~\cite{openrr1k}. 
All images are resized to $384 \times 384$ during training.
The model is implemented in PyTorch and optimized using Adam 
with learning rate $=10^{-4}$, weight decay $=0$, and batch size $=1$.
We apply CosineAnnealingLR scheduling ($T_{\text{max}}=10$, $\eta_{\text{min}}=8 \times 10^{-6}$) 
and train for 200 epochs.
For LFSB, we set window size $W = 12$, number of attention heads $H = \{2, 4, 8, 8, 8\}$ 
at decoder levels $\{0, 2, 3, 4, 5\}$ respectively, 
and the differential warmup period $E_{\text{warmup}} = 30$ epochs.
All experiments are conducted on a NVIDIA RTX 4090 GPU and are based on their original settings.

\begin{table}
    \centering
    \caption{\textbf{Cross-dataset evaluation on OpenRR-1K~\cite{openrr1k}.} 
    ReflexSplit achieves highest SSIM and lowest LPIPS, demonstrating superior perceptual quality and generalization capability.}
    \label{tab:openrr_results}
    \scalebox{0.75}{
    \begin{tabular}{lccccc}
        \toprule
        \textbf{Methods} & \textbf{PSNR}$\uparrow$ & \textbf{SSIM}$\uparrow$ & \textbf{LPIPS}$\downarrow$ & \textbf{NIQE}$\downarrow$ & \textbf{DISTS}$\downarrow$ \\
        \midrule
        DSRNet~\cite{hu2023single} & 24.5228 & 0.9139 & 0.1350 & \cellcolor{yellow!25}3.0109 & 0.0865 \\ 
        MaxRF~\cite{zhu2024revisiting} & 24.3146 & 0.9056 & 0.1180 & 3.2317 & 0.0750\\
        DExNet~\cite{huang2025lightweight} & \cellcolor{yellow!25}26.1434 & \cellcolor{orange!25}0.9351 & 0.1189 & 3.1006 & 0.0766\\
        DSIT~\cite{hu2024single} & \cellcolor{red!25}26.6672 & \cellcolor{yellow!25}0.9341 & \cellcolor{orange!25}0.1137 & \cellcolor{orange!25}2.9872 & \cellcolor{orange!25}0.0700 \\
        RDNet~\cite{zhao2024reversible} & 24.9297 & 0.9271 & \cellcolor{yellow!25}0.1140 & 3.1906 & \cellcolor{yellow!25}0.0728 \\
        \midrule
        \textbf{ReflexSplit (Ours)} & \cellcolor{orange!25}26.5824 & \cellcolor{red!25}\textbf{0.9372} & \cellcolor{red!25}\textbf{0.1087} & \cellcolor{red!25}\textbf{2.9764} & \cellcolor{red!25}\textbf{0.0684} \\
        \bottomrule
    \end{tabular}}
    \vspace{-0.4cm}
\end{table}

\subsection{Quantitative Comparison}
\label{sec:quantitative}

\begin{figure*}
    \centering
    \includegraphics[width=0.9\textwidth]{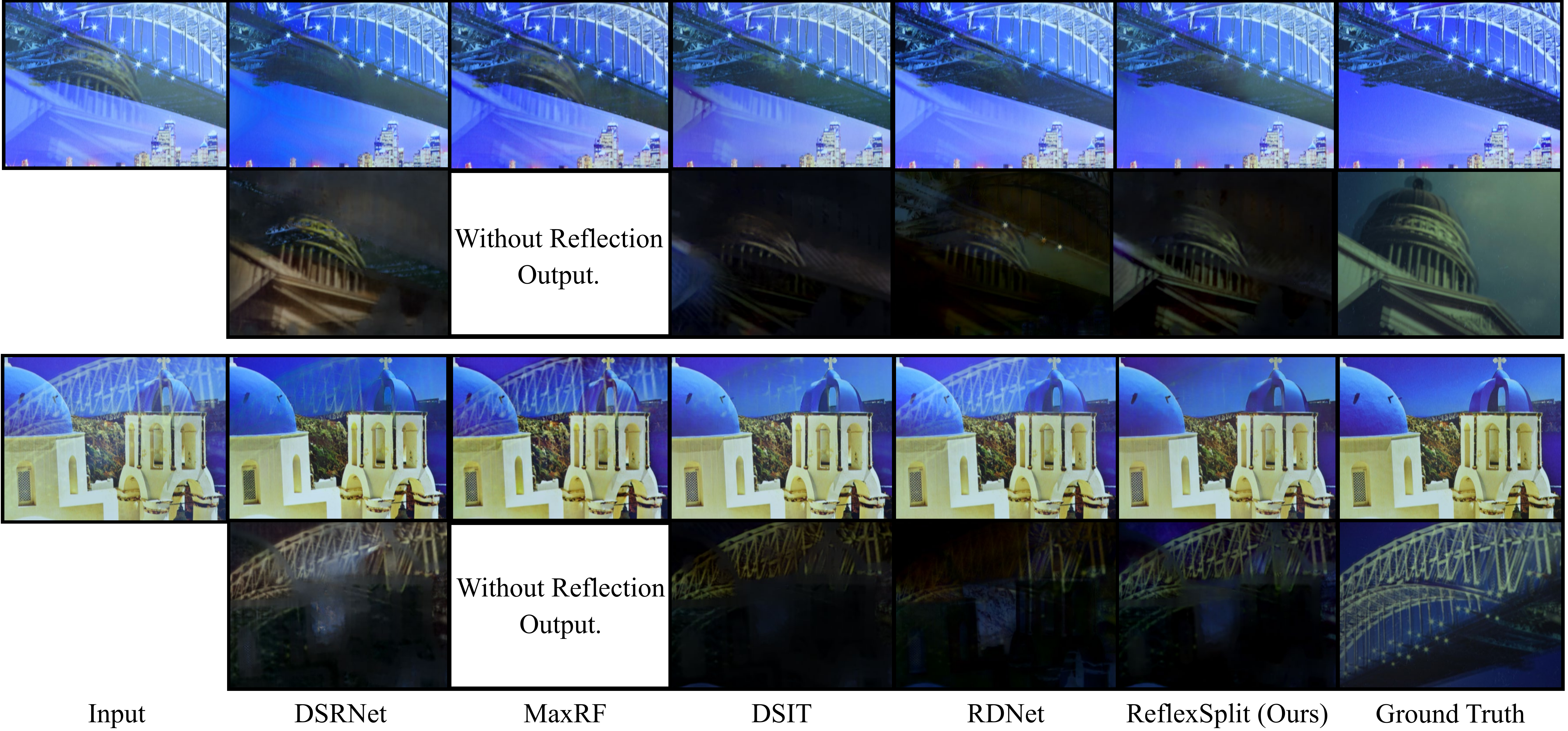}
    \caption{\textbf{Qualitative comparison on Postcard~\cite{wan2017benchmarking}.} 
    ReflexSplit achieves superior reflection separation and detail preservation.}
    \label{fig:postcard_comparison_main}
\vspace{-0.7mm}
\end{figure*}

We evaluate ReflexSplit against eleven state-of-the-art methods: 
Zhang et al.~\cite{zhang2018single}, 
ERRNet~\cite{wei2019single}, 
IBCLN~\cite{li2020single}, 
Zheng et al.~\cite{zheng2021single}, 
YTMT~\cite{hu2021trash}, 
DSRNet~\cite{hu2023single}, 
Zhong et al.~\cite{zhong2024language}, 
MaxRF~\cite{zhu2024revisiting}, 
DSIT~\cite{hu2024single}, 
DExNet~\cite{huang2025lightweight}, and RDNet~\cite{zhao2024reversible} 
on synthetic and real-world benchmarks.

\noindent\textbf{Synthetic Benchmarks.}
Table~\ref{tab:sirs_results} presents quantitative results across five synthetic datasets. The consistent SSIM leadership demonstrates ReflexSplit's superior perceptual quality through explicit layer separation, while competitive PSNR scores validate reconstruction fidelity. Notably, our method achieves this with 174M parameters—34\% fewer than RDNet and 28\% more than DSIT, offering an effective balance between performance and efficiency.

\noindent\textbf{Real-World Benchmarks and Cross-Dataset Generalization.}
On OpenRR-1K (Table~\ref{tab:openrr_results}), we evaluate cross-dataset generalization by directly testing models trained on PASCAL VOC~\cite{everingham2010pascal}, Real20, and Nature without fine-tuning.
We adopt both full-reference (PSNR/SSIM/LPIPS) and no-reference metrics (NIQE/DISTS) for comprehensive quality assessment.
ReflexSplit achieves the highest SSIM and lowest LPIPS, demonstrating superior perceptual quality and structural fidelity on unseen real-world data.
While DSIT obtains marginally higher PSNR, our method excels in perceptual metrics, indicating better preservation of fine-grained details and natural texture without over-smoothing.

\subsection{Qualitative Comparison}
We compare ReflexSplit against DSRNet~\cite{hu2023single}, MaxRF~\cite{zhu2024revisiting}, DSIT~\cite{hu2024single}, and RDNet~\cite{zhao2024reversible} on Postcard, Wild, and SolidObject datasets.

\begin{figure*}
    \centering
    \includegraphics[width=0.9\textwidth]{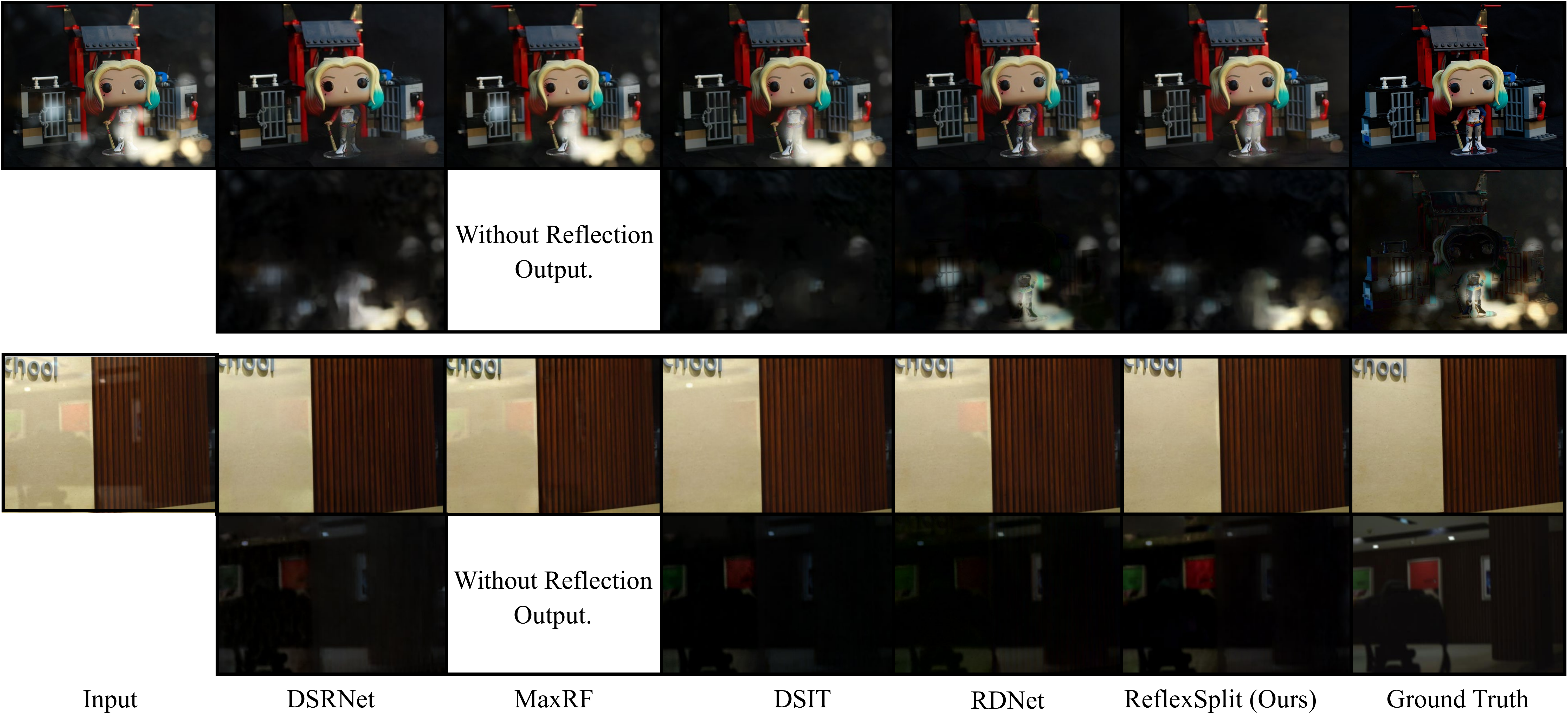}
    \caption{\textbf{Qualitative comparison on Real20 and Wild~\cite{wan2017benchmarking}.} ReflexSplit achieves superior reflection suppression while preserving fine-grained textures and color fidelity. Competing methods exhibit residual reflections, over-smoothing, or color distortion.}
    \label{fig:wild_comparison_main}
    \vspace{-0.4cm}
\end{figure*}

As shown in Figures~\ref{fig:postcard_comparison_main}--\ref{fig:wild_comparison_main}, competing methods exhibit distinct limitations: DSIT suffers transmission-reflection confusion; DSRNet fails in high-frequency regions; RDNet introduces color distortion; MaxRF loses edge details. ReflexSplit produces cleaner separation across diverse scenarios, preserving fine-grained textures, boundaries, and color fidelity. This results from CrGF maintaining hierarchical consistency through adaptive aggregation and LFSB preventing feature confusion via explicit fusion-separation.

\subsection{Ablation Study}
\label{sec:ablation}
We conduct ablation studies to validate each component's contribution using identical training settings described in Section~\ref{sec:experimentSetup}. All variants are trained for 200 epochs on the composite dataset and evaluated on Real20 and SIR² benchmarks.

\begin{figure}
    \centering
    \includegraphics[width=0.45\textwidth]{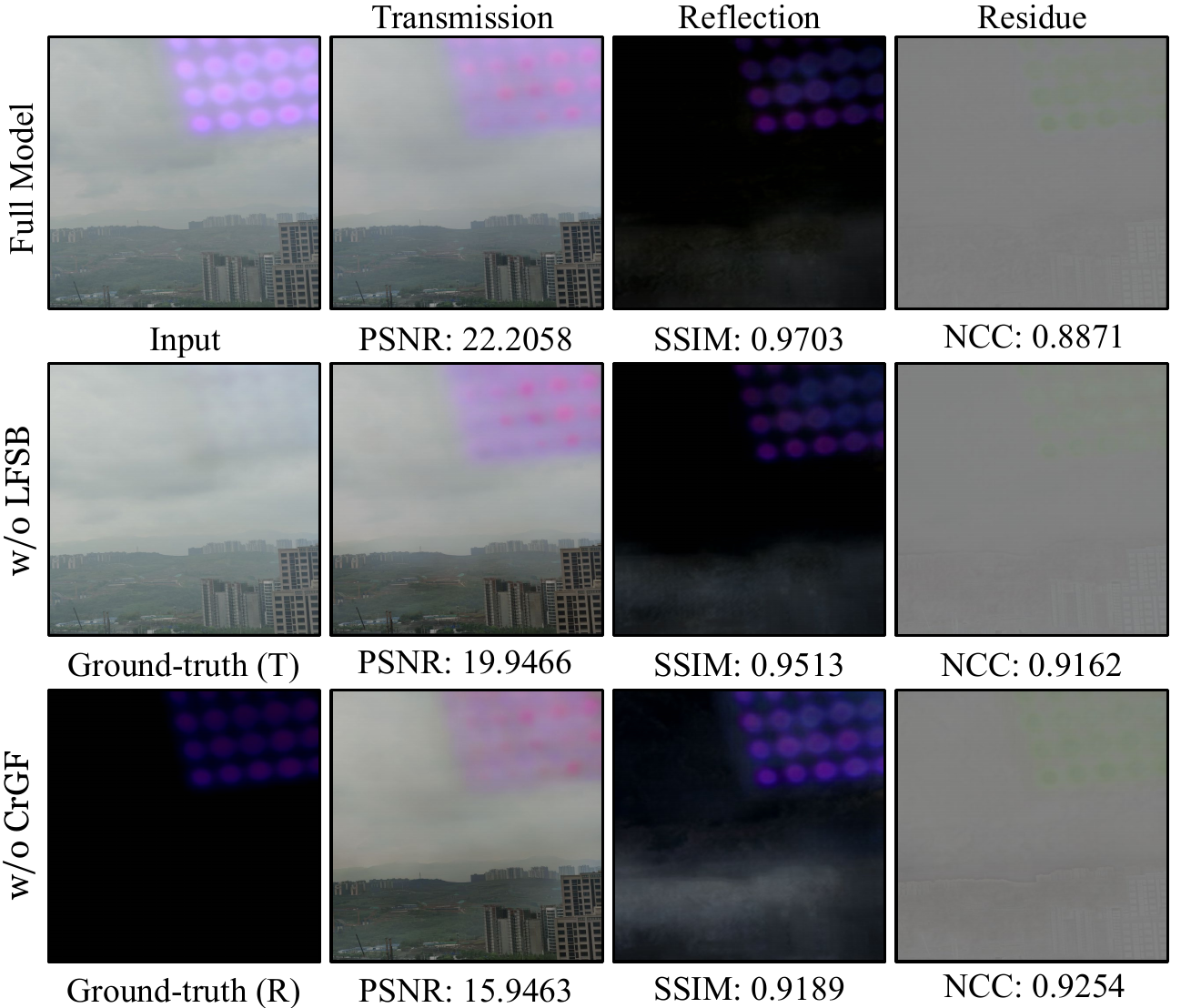}
    \caption{
    \textbf{Ablation study on challenging example.}
    Removing CrGF causes catastrophic failure, while removing LFSB leads to layer confusion. Both are essential for stable separation.
}
    \label{fig:ablation}
    \vspace{-0.4cm}
\end{figure}
\noindent\textbf{Effect of CrGF.} 
CrGF is critical for hierarchical feature consistency. Table~\ref{tab:ablation_crgf} shows that CrGF substantially outperforms alternative fusion strategies—direct aggregation, concatenation, and element-wise addition—demonstrating the importance of bidirectional gating for adaptive multi-scale coordination. This validates that stable separation requires both cross-scale aggregation (CrGF) and layer disentanglement (LFSB).

\noindent\textbf{Effect of LFSB Components.} 
Table~\ref{tab:ablation_lfsb} and Figure~\ref{fig:ablation} validate each component's contribution through progressive ablation. 
The baseline dual-stream decoder without explicit interaction suffers from severe transmission-reflection confusion. 
Bidirectional cross-stream projection in early fusion substantially improves performance by aligning semantic spaces, establishing a foundation for effective interaction.
Building upon early fusion, self-attention captures intra-layer spatial correlations independently for each stream, 
while cross-attention enables explicit inter-layer information exchange. 
However, simply aggregating SA and CA outputs risks reinforcing shared patterns rather than separating layers. 
The key innovation is introducing differential operators ($\mathbf{A}^t - \lambda\mathbf{A}^r$) that actively suppress cross-stream interference, 
preventing progressive feature entanglement. 
As shown in Figure~\ref{fig:ablation}, removing LFSB causes PSNR to drop to 19.946 and Normalized Cross-Correlation (NCC, measuring inter-stream similarity where lower is better) to 0.9254, confirming its critical role in layer separation.

\noindent\textbf{Effect of Training Strategy.}
Table~\ref{tab:ablation_training} validates our curriculum training: fixed strength causes instability, single-stage scheduling underperforms, while our full strategy combining both achieves optimal convergence by progressively adapting across spatial and temporal dimensions.
\begin{table}[t]
\centering
\caption{\textbf{Ablation study on CrGF.} We compare our Cross-scale Gated Fusion against alternative fusion strategies. All experiments use identical training settings.}
\vspace{-0.2mm}
\label{tab:ablation_crgf}
\resizebox{0.95\linewidth}{!}{
\begin{tabular}{lcccccc}
\toprule
\multirow{2}{*}{Fusion Strategy} & \multicolumn{2}{c}{Real20} & \multicolumn{2}{c}{SIR²} & \multicolumn{2}{c}{Nature} \\
\cmidrule(lr){2-3} \cmidrule(lr){4-5} \cmidrule(lr){6-7}
& PSNR↑ & SSIM↑ & PSNR↑ & SSIM↑ & PSNR↑ & SSIM↑ \\
\midrule
(a) No Fusion (Direct Aggregation) & 24.01 & 0.838 & 25.32 & 0.881 & 25.84 & 0.832 \\
(b) Simple Concatenation & 24.89 & 0.828 & 25.67 & 0.878 & 26.21 & 0.841 \\
(c) Element-wise Addition & 25.01 & 0.830 & 25.78 & 0.883 & 26.45 & 0.846 \\
(d) CrGF (Ours) & \textbf{25.22} & \textbf{0.846} & \textbf{26.33} & \textbf{0.896} & \textbf{27.03} & \textbf{0.854} \\
\bottomrule
\end{tabular}
}
\vspace{-0.5mm}
\end{table}

\begin{table}[t]
\centering
\caption{\textbf{Ablation study on LFSB components.} We progressively add each component to evaluate its contribution.}
\label{tab:ablation_lfsb}
\resizebox{0.85\linewidth}{!}{
\begin{tabular}{lccccccc}
\toprule
\multirow{2}{*}{Variant} & \multirow{2}{*}{\shortstack{Early\\Fusion}} & \multirow{2}{*}{\shortstack{SA\\Only}} & \multirow{2}{*}{\shortstack{SA+CA}} & \multirow{2}{*}{\shortstack{Diff\\Sep}} & \multirow{2}{*}{\shortstack{Late\\Fusion}} & \multicolumn{2}{c}{Real20} \\
\cmidrule(lr){7-8}
& & & & & & PSNR↑ & SSIM↑ \\
\midrule
(a) Baseline & \xmark & \xmark & \xmark & \xmark & \xmark & 23.87 & 0.812 \\
(b) + Early Fusion & \cmark & \xmark & \xmark & \xmark & \xmark & 24.32 & 0.823  \\
(c) + SA & \cmark & \cmark & \xmark & \xmark & \xmark & 24.61 & 0.831  \\
(d) + SA + CA & \cmark & \xmark & \cmark & \xmark & \xmark & 24.89 & 0.838  \\
(e) + Diff Sep & \cmark & \xmark & \cmark & \cmark & \xmark & 25.08 & 0.842  \\
(f) Full LFSB (Ours) & \cmark & \xmark & \cmark & \cmark & \cmark & \textbf{25.22} & \textbf{0.846} \\
\bottomrule
\end{tabular}
} 
\vspace{-0.5mm}
\end{table}
\begin{table}[t]
\centering
\caption{
\textbf{Ablation on training strategy.}
The full curriculum strategy achieves the best performance, confirming the effectiveness of both depth-dependent initialization and epoch-wise warmup.
}
\label{tab:ablation_training}
\scalebox{0.65}{
\begin{tabular}{lcccc}
\toprule
\multirow{2}{*}{Strategy} & \multicolumn{2}{c}{Real20} & \multicolumn{2}{c}{SIR$^2$} \\
\cmidrule(lr){2-3} \cmidrule(lr){4-5}
& PSNR↑ & SSIM↑ & PSNR↑ & SSIM↑ \\
\midrule
Fixed $\lambda$ ($=0.5$) & 24.83 & 0.821 & 25.89 & 0.883 \\
Warmup only & 24.95 & 0.835 & 26.11 & 0.891 \\
Depth-init only & 25.08 & 0.840 & 26.18 & 0.889  \\
\textbf{Full strategy (Ours)} & \textbf{25.22} & \textbf{0.846} & \textbf{26.33} & \textbf{0.896}  \\
\bottomrule
\end{tabular}
}
\vspace{-0.4cm}
\end{table}
\section{Conclusion}
\label{sec:conclusion}
We present ReflexSplit, a dual-stream framework that addresses transmission-reflection confusion in SIRS through explicit layer fusion-separation. Our approach integrates three key innovations: Cross-scale Gated Fusion (CrGF) adaptively aggregates multi-scale features across decoder depths, stabilizing gradient flow and maintaining hierarchical consistency; Layer Fusion-Separation Blocks (LFSB) alternate between fusion for shared structure extraction and differential separation for layer-specific disentanglement via cross-stream attention subtraction; curriculum training progressively strengthens differential separation through depth-dependent initialization and epoch-wise warmup. This synergistic design prevents progressive feature entanglement that plagues existing methods. Extensive experiments on synthetic and real-world benchmarks demonstrate state-of-the-art performance with superior perceptual quality and robust generalization.

\clearpage
\setcounter{page}{1}
\maketitlesupplementary

\section*{Overview}
This supplementary material provides comprehensive details to support the main paper. The document is organized as follows:
\begin{itemize}
    \item \textbf{Section 1: Algorithm Description} -- Complete algorithmic specification of the ReflexSplit training pipeline, including dual-branch feature extraction, hierarchical decoding with CrGF and LFSB, and curriculum learning strategy.
    
    \item \textbf{Section 2: Loss Functions Details} -- Detailed formulation of all loss components: reconstruction losses, perceptual loss, exclusion loss, reconstruction consistency, and color consistency loss with weight configurations.
    
    \item \textbf{Section 3: Feature Separation Analysis with DSIT} -- t-SNE and PCA analysis comparing progressive disentanglement strategy of ReflexSplit against immediate separation approach of DSIT, demonstrating more distributed feature representations and curriculum-based layer learning.
    
    \item \textbf{Section 4: Comparison with RDNet on Real-World Scenarios} -- Performance comparison with RDNet on OpenRR-1K, showing ReflexSplit achieves improvements on 63\% of test images. Discussion of complementary architectural philosophies between reversible designs and explicit layer separation approaches.
    
    \item \textbf{Section 5: Additional Visual Comparisons} -- Extensive qualitative results on OpenRR-1K and SIR$^2$ datasets, demonstrating reflection suppression, color fidelity, and detail preservation compared to state-of-the-art methods.
    
    \item \textbf{Section 6: Failure Cases and Limitations} -- Analysis of challenging scenarios where ReflexSplit struggles: complex outdoor lighting, specular reflections, and mixed indoor-outdoor scenes with extreme brightness differences.
    
    \item \textbf{Section 7: Network Architecture Details} -- Complete architecture specification with layer-by-layer breakdown of dual-branch encoders, feature mixing, hierarchical decoder stages, and output generation modules.

    \item \textbf{Section 8: More Complexity Comparison} -- We present a comprehensive efficiency analysis comparing ReflexSplit with existing methods, demonstrating its practical advantage as a single-stage solution with 
competitive computational cost.
\end{itemize}

\vspace{5mm}

\section{Algorithm Description}
\label{sec:algorithm}
We provide a detailed algorithmic description of ReflexSplit in Algorithm~\ref{alg:reflexsplit}, which illustrates the complete training pipeline including dual-branch feature extraction, hierarchical decoding with CrGF and LFSB, and curriculum learning strategy.

\begin{algorithm}[t]
\caption{ReflexSplit Training Algorithm}
\label{alg:reflexsplit}
\begin{algorithmic}[1]
\footnotesize
\REQUIRE Mixed image $\mathbf{I} \in \mathbb{R}^{H \times W \times 3}$, ground truth transmission $\mathbf{T}$ and reflection $\mathbf{R}$, training epoch $e$, warmup epoch $E_{\text{warmup}}$
\ENSURE Predicted transmission $\hat{\mathbf{T}}$, reflection $\hat{\mathbf{R}}$, and residual $\hat{\mathbf{RR}}$

\STATE \textbf{// Stage 1: Dual-Branch Feature Extraction}
\STATE Extract global semantic priors via Swin Transformer:
\STATE \quad $\{\mathbf{P}_2, \mathbf{P}_3, \mathbf{P}_4, \mathbf{P}_5\} \leftarrow \text{GFEB}(\mathbf{I})$
\STATE Extract local texture features via MuGI-CNN:
\STATE \quad $\{\mathbf{E}_0, \mathbf{E}_1, \mathbf{E}_2, \mathbf{E}_3, \mathbf{E}_4, \mathbf{E}_5\} \leftarrow \text{LFEB}(\mathbf{I})$

\STATE \textbf{// Stage 2: Hierarchical Decoding with CrGF and LFSB}
\FOR{decoder level $\ell = 5$ to $0$}
    \IF{$\ell \in \{4, 3, 2\}$}
        \STATE \textbf{// Cross-scale Gated Fusion (CrGF)}
        \STATE $\mathbf{F}^{\text{raw}}_{\ell} \leftarrow \mathbf{F}_{\ell+1} + \mathbf{P}_{\ell} + \mathbf{E}_{\ell}$
        \STATE Compute bidirectional gating:
        \STATE \quad $\mathbf{F}^{\text{main}}_{\ell} = \mathcal{G}_1(\mathbf{F}^{\text{raw}}_{\ell}) \odot \mathcal{G}_2(\mathbf{F}_{\ell+1})$
        \STATE \quad $\mathbf{F}^{\text{aux}}_{\ell} = \mathcal{G}_1(\mathbf{F}_{\ell+1}) \odot \mathcal{G}_2(\mathbf{F}^{\text{raw}}_{\ell})$
        \STATE $\mathbf{F}^{\text{fused}}_{\ell} = w^{(1)}_{\ell}\phi_1(\mathbf{F}^{\text{main}}_{\ell}) + w^{(2)}_{\ell}\phi_2(\mathbf{F}^{\text{aux}}_{\ell})$
    \ELSE
        \STATE \textbf{// Direct Aggregation}
        \STATE $\mathbf{F}^{\text{fused}}_{\ell} \leftarrow \mathbf{F}_{\ell+1} + \mathbf{E}_{\ell}$
    \ENDIF
    
    \STATE \textbf{// Layer Fusion-Separation Block (LFSB)}
    \STATE \textcolor{blue}{/* Early Fusion: Bidirectional Projection */}
    \STATE $\mathbf{F}^{t'}_{\ell} = \mathbf{W}^t[\mathbf{F}^t_{\ell} \| \mathbf{F}^r_{\ell}]$
    \STATE $\mathbf{F}^{r'}_{\ell} = \mathbf{W}^r[\mathbf{F}^r_{\ell} \| \mathbf{F}^t_{\ell}]$
    
    \STATE \textcolor{blue}{/* Differential Dual-Dimensional Attention */}
    \STATE Compute self-attention (batch-wise):
    \STATE \quad $\mathbf{A}^t_{\text{SA}}, \mathbf{A}^r_{\text{SA}} \leftarrow \text{SA}(\text{concat}([\mathbf{F}^{t'}_{\ell}, \mathbf{F}^{r'}_{\ell}], \text{dim}=0))$
    \STATE Compute cross-attention (sequence-wise):
    \STATE \quad $\mathbf{A}^t_{\text{CA}}, \mathbf{A}^r_{\text{CA}} \leftarrow \text{CA}(\text{concat}([\mathbf{F}^{t'}_{\ell}, \mathbf{F}^{r'}_{\ell}], \text{dim}=1))$
    
    \STATE \textcolor{blue}{/* Differential Separation with Curriculum */}
    \STATE Compute depth-dependent weight:
    \STATE \quad $\lambda^{\text{init}}_{\ell} = 0.8 - 0.6 \exp(-0.3\ell)$
    \STATE Compute epoch-wise warmup:
    \STATE \quad $\lambda_{\text{diff}}(e) = \begin{cases} 0.1 + 0.9\frac{e}{E_{\text{warmup}}}, & e < E_{\text{warmup}}\\ 1.0, & e \geq E_{\text{warmup}} \end{cases}$
    \STATE $\lambda_{\ell}(e) \leftarrow \lambda^{\text{init}}_{\ell} \cdot \lambda_{\text{diff}}(e)$
    \STATE Apply differential operators:
    \STATE \quad $\mathbf{A}^t_{\text{diff}} = (\mathbf{A}^t_{\text{SA}} + \mathbf{A}^t_{\text{CA}}) - \sigma(\lambda_{\ell})(\mathbf{A}^r_{\text{SA}} + \mathbf{A}^r_{\text{CA}})$
    \STATE \quad $\mathbf{A}^r_{\text{diff}} = (\mathbf{A}^r_{\text{SA}} + \mathbf{A}^r_{\text{CA}} ) - \sigma(\lambda_{\ell})(\mathbf{A}^t_{\text{SA}} + \mathbf{A}^t_{\text{CA}})$
    
    \STATE \textcolor{blue}{/* Late Fusion: FFN with Residual */}
    \STATE $\mathbf{F}^t_{\ell+1} = \mathbf{F}^t_{\ell} + \text{FFN}(\mathbf{A}^t_{\text{diff}})$
    \STATE $\mathbf{F}^r_{\ell+1} = \mathbf{F}^r_{\ell} + \text{FFN}(\mathbf{A}^r_{\text{diff}})$
\ENDFOR

\STATE \textbf{// Stage 3: Output Generation}
\STATE $\hat{\mathbf{T}}, \hat{\mathbf{R}} \leftarrow \text{Conv}_{3 \times 3}(\mathbf{F}^t_0), \text{Conv}_{3 \times 3}(\mathbf{F}^r_0)$
\STATE $\hat{\mathbf{RR}} \leftarrow \text{LRM}(\mathbf{F}^t_0 + \mathbf{F}^r_0)$

\STATE \textbf{// Loss Computation}
\STATE $\mathcal{L}_{\text{total}} = \lambda_{\text{rec}}\mathcal{L}_{\text{rec}} + \lambda_{\text{refl}}\mathcal{L}_{\text{refl}} + \lambda_{\text{vgg}}\mathcal{L}_{\text{vgg}}$
\STATE \quad $+ \lambda_{\text{exclu}}\mathcal{L}_{\text{exclu}} + \lambda_{\text{recons}}\mathcal{L}_{\text{recons}}+ \lambda_{\text{color}}\mathcal{L}_{\text{color}}$
\RETURN $\hat{\mathbf{T}}, \hat{\mathbf{R}}, \hat{\mathbf{RR}}$
\end{algorithmic}
\end{algorithm}

\section{Loss Functions Details}
Let $\hat{\mathbf{T}}$, $\hat{\mathbf{R}}$, and $\hat{\mathbf{RR}}$ denote the predicted transmission, reflection, and residual layers, respectively, with $\mathbf{T}$ and $\mathbf{R}$ as ground truth. Our training objective combines multiple complementary loss terms:

\begin{equation}
\begin{aligned}
\mathcal{L}_{\text{total}} 
&= \lambda_\text{rec}\,\mathcal{L}_{\text{rec}} 
+ \lambda_\text{refl}\,\mathcal{L}_{\text{refl}} 
+ \lambda_\text{vgg}\,\mathcal{L}_{\text{vgg}} \\
&\quad + \lambda_\text{exclu}\,\mathcal{L}_{\text{exclu}}
+ \lambda_\text{recons}\,\mathcal{L}_{\text{recons}} + \lambda_\text{color}\,\mathcal{L}_{\text{color}}.
\end{aligned}
\end{equation}

\noindent\textbf{Reconstruction losses.} 
For transmission layer reconstruction, we employ Charbonnier loss~\cite{charbonier} to handle outliers robustly:
\begin{equation}
\mathcal{L}_{\text{rec}} = \sqrt{\|\hat{\mathbf{T}} - \mathbf{T}\|^2 + \epsilon^2}, \quad \epsilon = 10^{-3}.
\end{equation}
For reflection layer supervision, we use $\ell_1$ loss:
\begin{equation}
\mathcal{L}_{\text{refl}} = \|\hat{\mathbf{R}} - \mathbf{R}\|_1.
\end{equation}

\noindent\textbf{Perceptual loss.} 
To preserve semantic content and texture details, we adopt VGG perceptual loss~\cite{vgg} using features extracted from layers $\{2, 7, 12, 21, 30\}$ of a pretrained VGG-19 network:
\begin{equation}
\mathcal{L}_{\text{vgg}} = \sum_{i \in \{2,7,12,21,30\}} w_i \cdot \|\phi_i(\hat{\mathbf{T}}) - \phi_i(\mathbf{T})\|_1,
\end{equation}
where $\phi_i(\cdot)$ denotes the feature extractor at layer $i$ and $w_i$ are weights balancing contributions across layers.

\noindent\textbf{Exclusion loss.} 
Following~\cite{zhang2018single}, we enforce gradient orthogonality between transmission and reflection to encourage layer independence:
\begin{equation}
\mathcal{L}_{\text{exclu}} = \sum_{l=1}^{3} \left( \|\nabla_x \hat{\mathbf{T}} \odot \nabla_x \hat{\mathbf{R}}\|_1 + \|\nabla_y \hat{\mathbf{T}} \odot \nabla_y \hat{\mathbf{R}}\|_1 \right),
\end{equation}
where $\nabla_x$ and $\nabla_y$ denote spatial gradients along horizontal and vertical directions, and $\odot$ represents element-wise multiplication. This term penalizes overlapping gradient patterns, thereby promoting structural separation between layers.

\noindent\textbf{Reconstruction consistency.} 
To ensure the decomposed layers reconstruct the input accurately, we enforce:
\begin{equation}
\mathcal{L}_{\text{recons}} = \|\hat{\mathbf{T}} + \hat{\mathbf{R}} + \hat{\mathbf{RR}} - \mathbf{I}\|_1,
\end{equation}
where $\mathbf{I}$ is the input mixed image. This constraint guarantees that no information is lost during decomposition.

\noindent\textbf{Color consistency loss.} 
To maintain color fidelity in the separated reflection layer, we introduce a color consistency term that matches color statistics between prediction and ground truth:
\begin{equation}
\mathcal{L}_{\text{color}} = \|\mu(\hat{\mathbf{R}}) - \mu(\mathbf{R})\|_1 + \|\sigma(\hat{\mathbf{R}}) - \sigma(\mathbf{R})\|_1,
\end{equation}
where $\mu(\cdot)$ and $\sigma(\cdot)$ compute channel-wise mean and standard deviation.

\noindent\textbf{Loss weights.} 
We set the loss weights as: $\lambda_\text{rec} = 1.0$, $\lambda_\text{refl} = 0.5$, $\lambda_\text{vgg} = 0.1$, $\lambda_\text{exclu} = 1.0$, $\lambda_\text{recons} = 0.2$, and $\lambda_\text{color} = 0.1$. 

\begin{figure*}[t]
\centering
\includegraphics[width=1\textwidth]{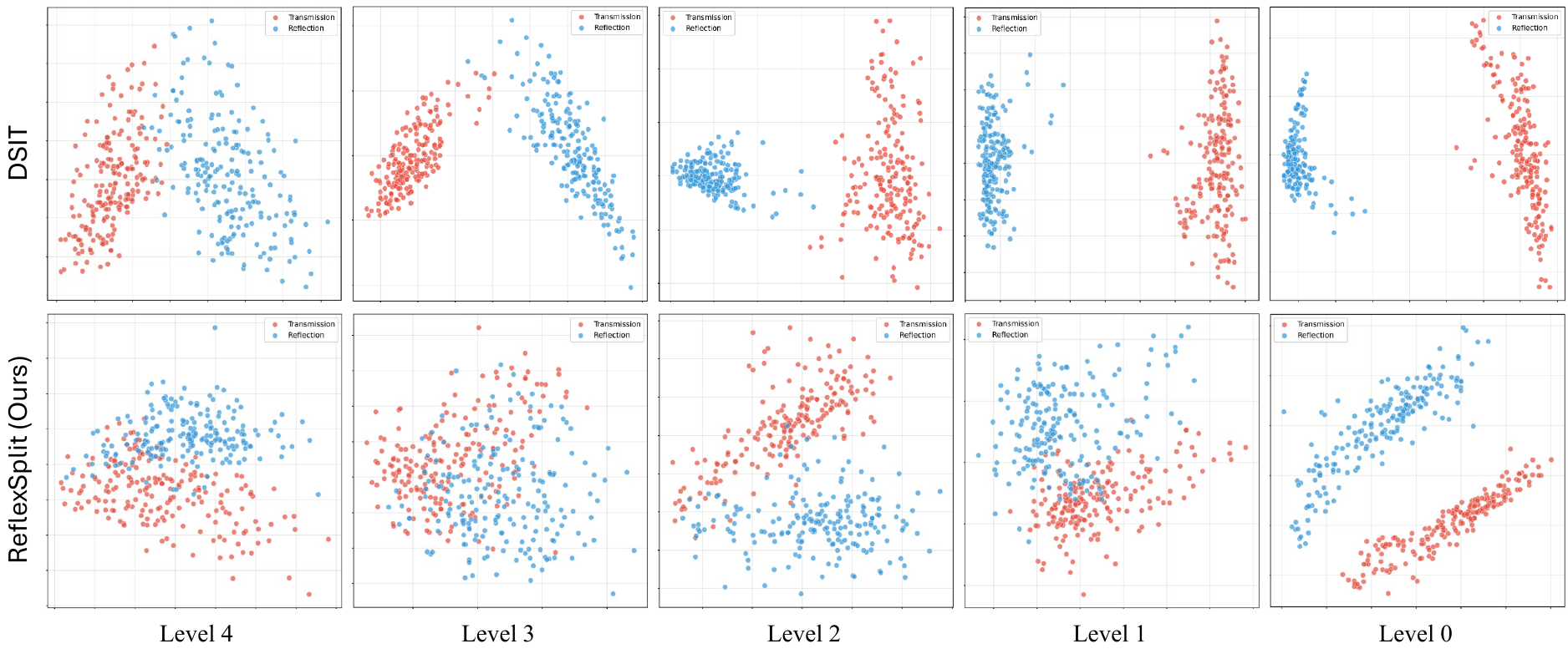}
\caption{\textbf{t-SNE visualization reveals different separation strategies.} DSIT~\cite{hu2024single} (top) maintains strong separation across all levels, while ReflexSplit (bottom) demonstrates progressive disentanglement from overlap (Level 4) to clear separation (Level 0). Red/blue: transmission/reflection features.}
\label{fig:TNSE}
\end{figure*}

\begin{figure}[t]
\centering
\includegraphics[width=0.48\textwidth]{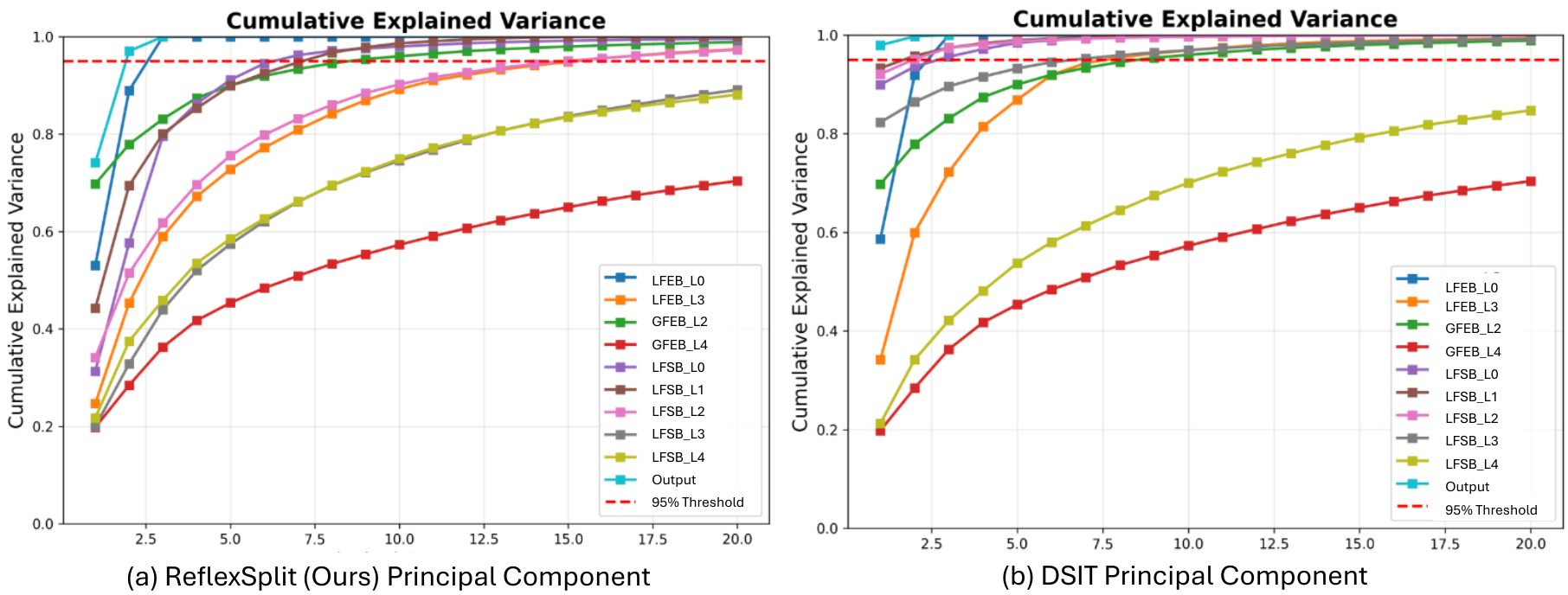}
\caption{\textbf{Cumulative explained variance comparison.} (a) ReflexSplit shows gradual accumulation indicating richer representations. (b) DSIT exhibits steeper accumulation with variance concentrated in fewer components.}
\label{fig:PCA}
\end{figure}

\section{Feature Separation Analysis: Comparison with DSIT}
\label{sec:dsit_analysis}

DSIT~\cite{hu2024single} represents a strong baseline with its dual-stream transformer architecture. However, we observe that different separation strategies may lead to distinct feature learning behaviors under complex real-world mixing conditions.

Figures~\ref{fig:TNSE} and~\ref{fig:PCA} provide feature space analysis comparing ReflexSplit and DSIT on OpenRR-1K~\cite{openrr1k}, revealing fundamental differences in separation strategies.

\noindent\textbf{t-SNE feature distribution.} 
Figure~\ref{fig:TNSE} visualizes transmission and reflection features across LFSB decoder levels. DSIT maintains strong separation throughout all levels, which represents an architectural philosophy that prioritizes immediate layer distinction. ReflexSplit demonstrates \textit{progressive disentanglement}: features overlap at Level 4 to facilitate shared feature learning at deep layers, then gradually separate toward Level 0. This curriculum-based strategy $\mathbf{A}^t - \lambda(e) \mathbf{A}^r$ may enable learning shared features before specializing to layer-specific characteristics.

\noindent\textbf{Principal component analysis.} 
Figure~\ref{fig:PCA} shows ReflexSplit features accumulate variance gradually, requiring more principal components to reach 95\% threshold—indicating more distributed representations. DSIT exhibits steeper accumulation with variance concentrated in fewer components, suggesting more compact encoding. While both achieve efficient output encoding, ReflexSplit's intermediate LFSB features require more principal components to capture equivalent variance, which may facilitate progressive layer disentanglement.

These findings validate our differential attention design with curriculum learning, demonstrating that progressive separation represents a viable alternative to immediate separation approaches, particularly for complex real-world reflection removal scenarios.

\begin{table}[t]
\centering
\caption{\textbf{Performance margin distribution on OpenRR-1K.} Comparison of ReflexSplit against RDNet~\cite{zhao2024reversible} on both test and validation splits. $\Delta_{\text{PSNR}} = \text{PSNR}_{\text{Ours}} - \text{PSNR}_{\text{RDNet}}$.}
\label{tab:margin_distribution_both}
\scalebox{0.88}{
\begin{tabular}{c|cc|cc}
\hline
\multirow{2}{*}{\textbf{Condition}} & \multicolumn{2}{c|}{\textbf{Test Set (99 images)}} & \multicolumn{2}{c}{\textbf{Val Set (100 images)}} \\
& \textbf{Count} & \textbf{Percentage} & \textbf{Count} & \textbf{Percentage} \\
\hline
$\Delta_{\text{PSNR}} > 0$ dB & 63 / 99 & 63.64\% & 62 / 100 & 62.00\% \\
$\Delta_{\text{PSNR}} > 3$ dB & 20 / 99 & 20.20\% & 18 / 100 & 18.00\% \\
$\Delta_{\text{PSNR}} > 5$ dB & 8 / 99 & 8.08\% & 10 / 100 & 10.00\% \\
$\Delta_{\text{PSNR}} > 7$ dB & 5 / 99 & 5.05\% & 4 / 100 & 4.00\% \\
\hline
\end{tabular}}
\end{table}

\section{Comparison with RDNet on Real-World Scenarios}
\label{sec:rdnet_analysis}

RDNet~\cite{zhao2024reversible} represents a strong baseline with its reversible architecture that preserves information flow through the network. While implicit feature learning through reversible architectures shows strong performance across diverse scenarios, we observe that explicit layer separation constraints may provide complementary advantages in certain complex cases, particularly scenes with over-exposure, specular highlights, and spatially-varying attenuation.

Table~\ref{tab:margin_distribution_both} provides a detailed performance margin distribution comparing ReflexSplit and RDNet on OpenRR-1K~\cite{openrr1k}. ReflexSplit achieves improvements on 63\% of test images, with notable gains ($>$3 dB) on 20\% of cases. These improvements are particularly evident in challenging scenarios shown in Figure~\ref{fig:openrr_additional}.

We hypothesize that these gains are related to our explicit differential attention mechanism $\mathbf{A}^t - \lambda \mathbf{A}^r$, which provides clear separation constraints at each decoder level. This architectural choice appears to offer advantages for maintaining layer distinction under complex nonlinear mixing, representing a complementary design philosophy to RDNet's reversible approach. Both strategies demonstrate merit, with performance differences dependent on specific scene characteristics.

\section{Additional Visual Comparisons}
\label{sec:additional_visual}

\begin{figure}[t]
    \centering
    \begin{subfigure}[b]{0.48\textwidth}
        \centering
        \includegraphics[width=\textwidth]{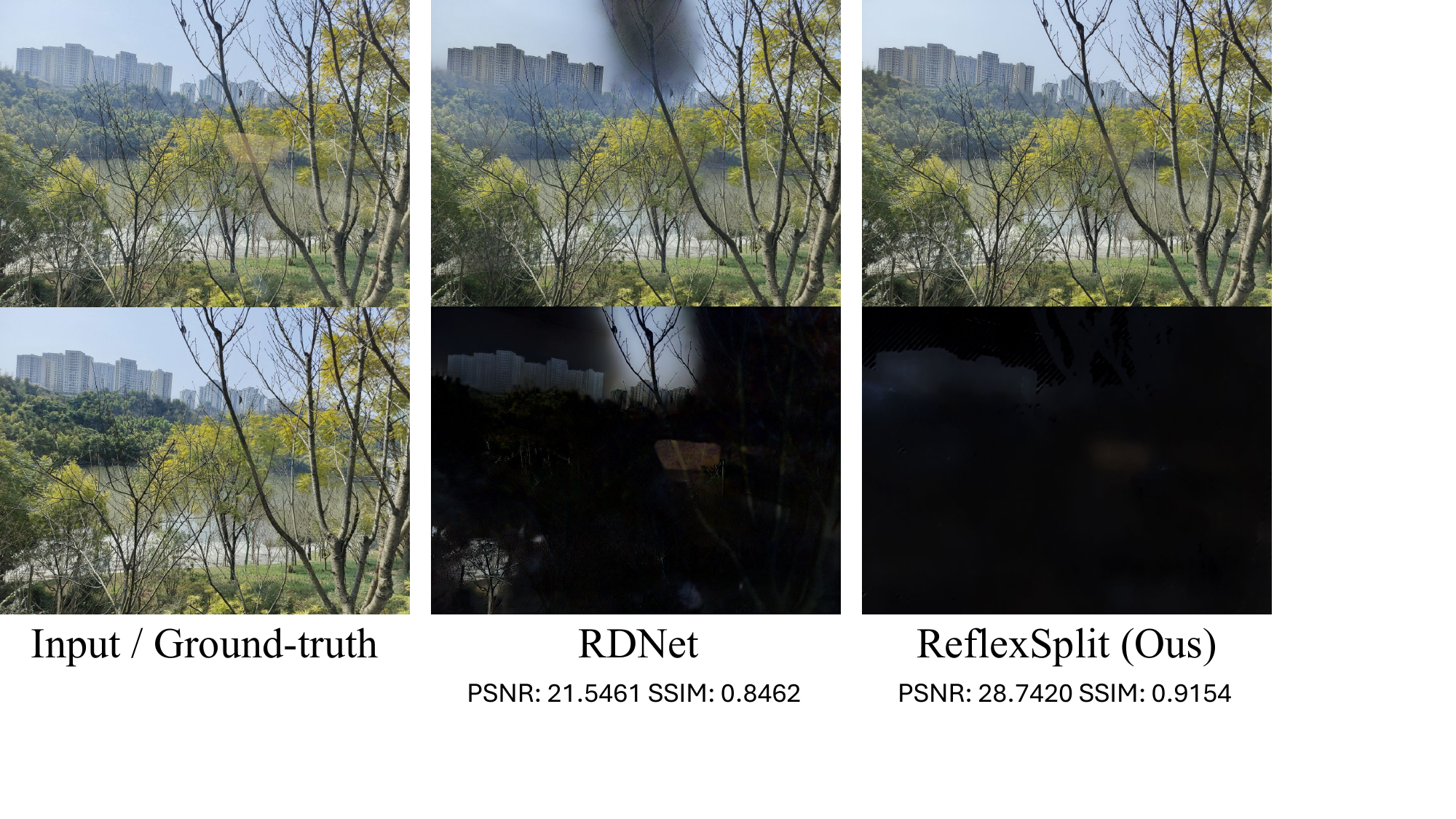}
        \label{fig:Reflex_sup1}
    \end{subfigure}
    \hfill
    \begin{subfigure}[b]{0.48\textwidth}
        \centering
        \includegraphics[width=\textwidth]{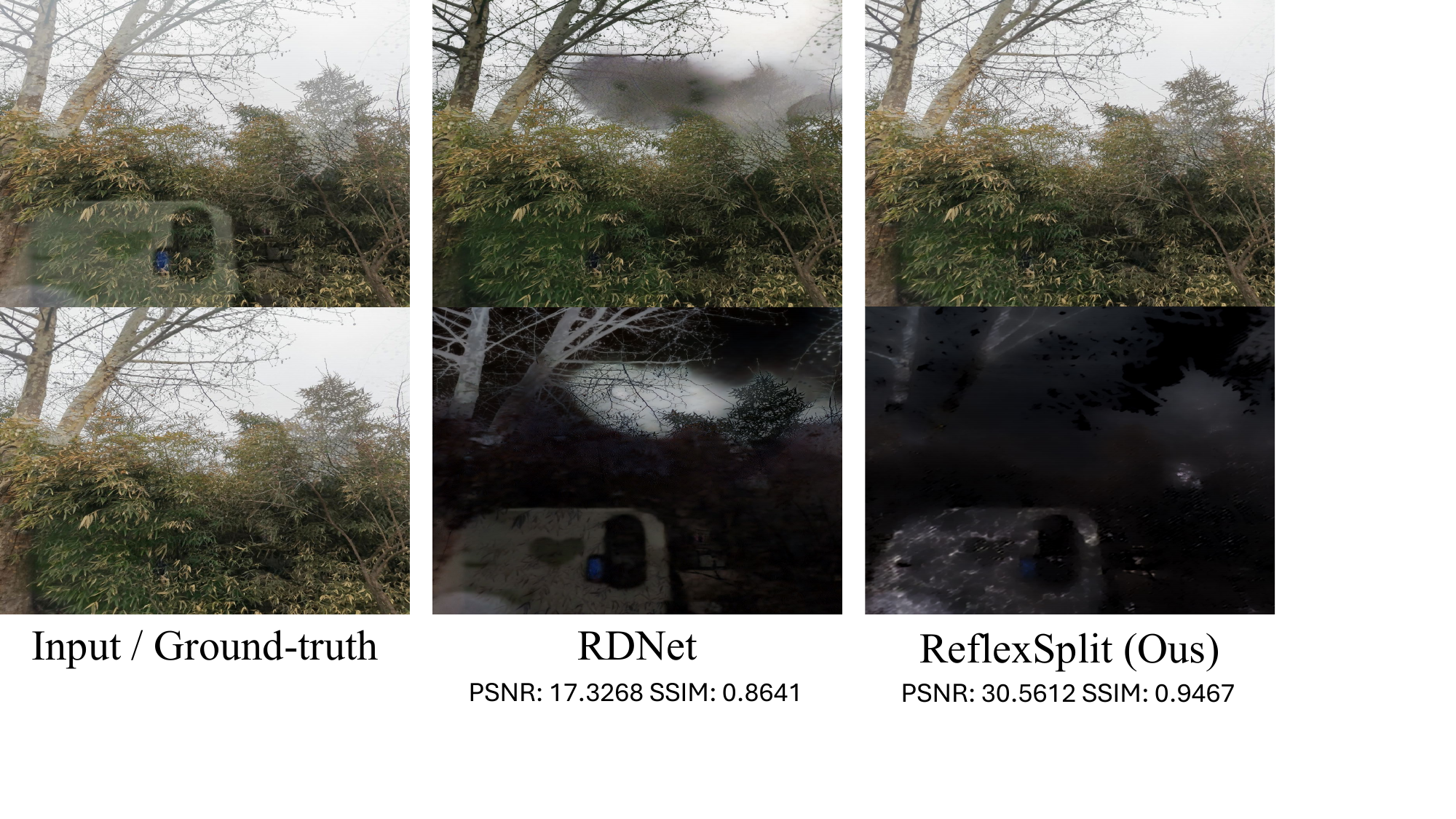}
        \label{fig:Reflex_sup2}
    \end{subfigure}
    
    \vspace{0.2cm}
    
    \begin{subfigure}[b]{0.48\textwidth}
        \centering
        \includegraphics[width=\textwidth]{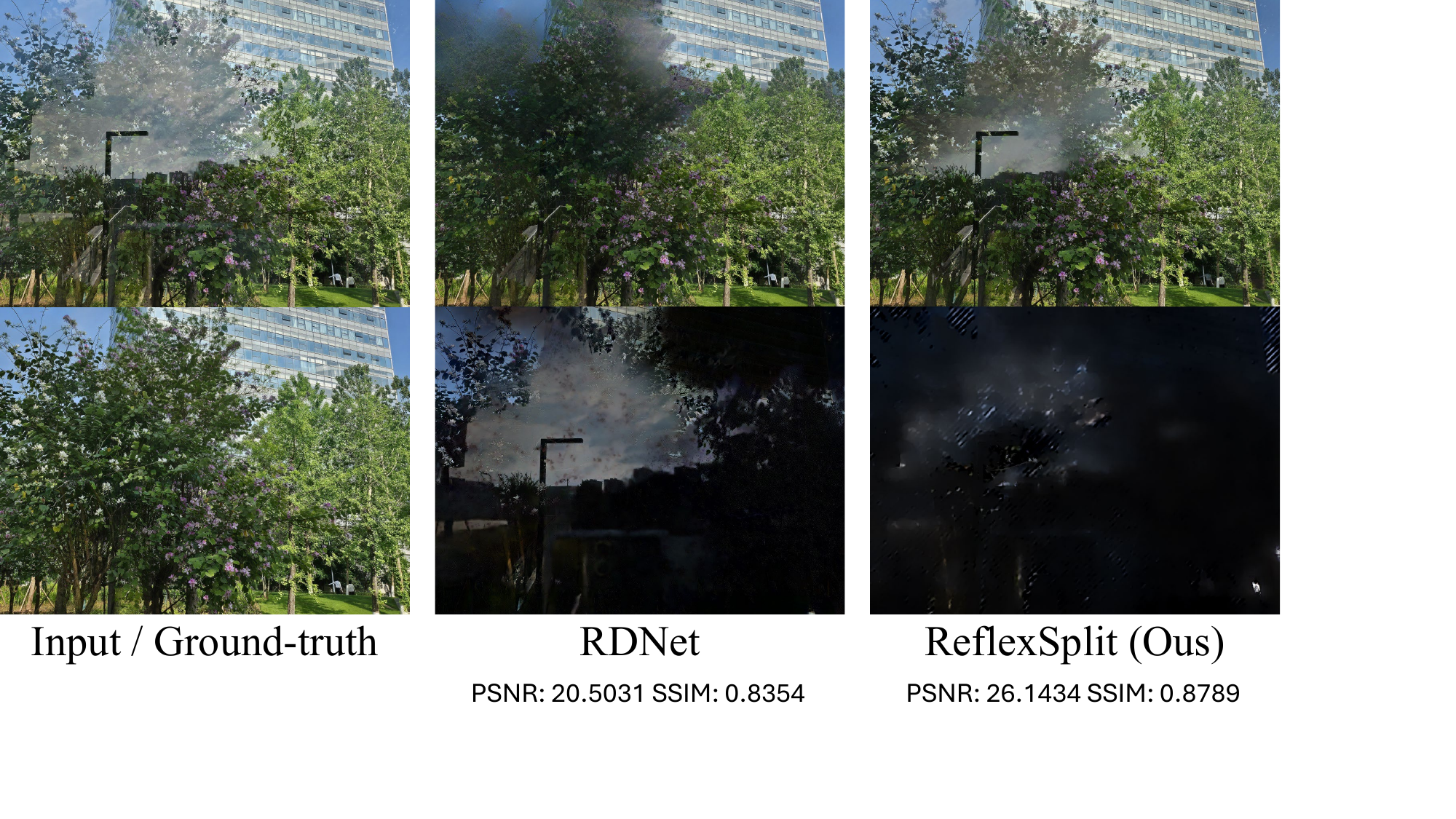}
        \label{fig:Reflex_sup3}
    \end{subfigure}
    
    \caption{\textbf{Challenging real-world scenarios on OpenRR-1K~\cite{openrr1k}.} 
    Comparison of ReflexSplit with state-of-the-art RDNet~\cite{zhao2024reversible} on diverse degradation types. Our differential attention mechanism provides explicit layer separation constraints, achieving natural color fidelity and sharp detail preservation in these challenging cases.}
    \label{fig:openrr_additional}
    \vspace{-0.2cm}
\end{figure}

\begin{figure*}
    \centering
    \includegraphics[width=\textwidth]{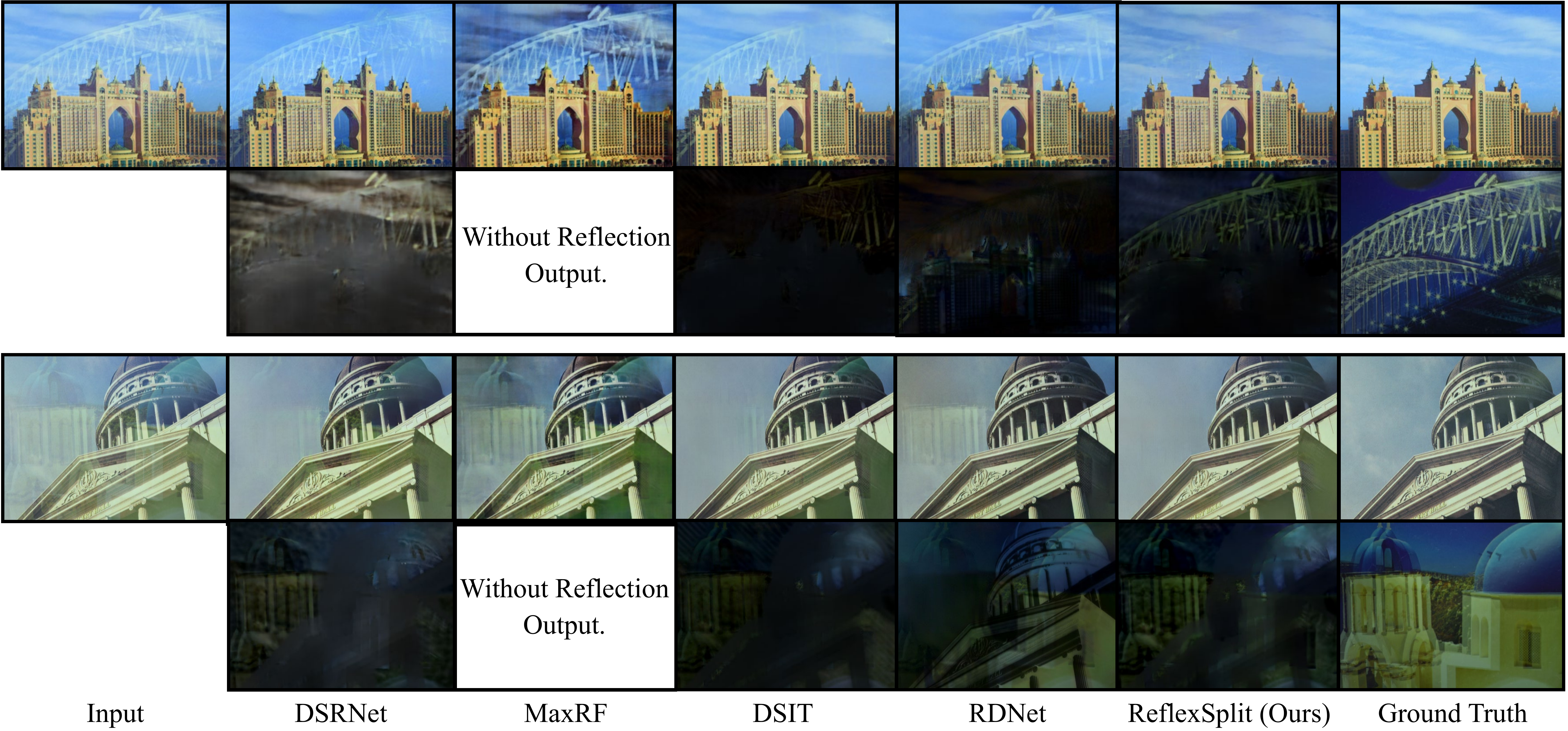}
    \caption{\textbf{Qualitative comparison on Postcard~\cite{wan2017benchmarking}.} 
    ReflexSplit achieves superior reflection suppression with faithful detail preservation.}
    \label{fig:postcard_comparison}
    \vspace{-0.2cm}
\end{figure*}

\begin{figure*}
    \centering
    \includegraphics[width=\textwidth]{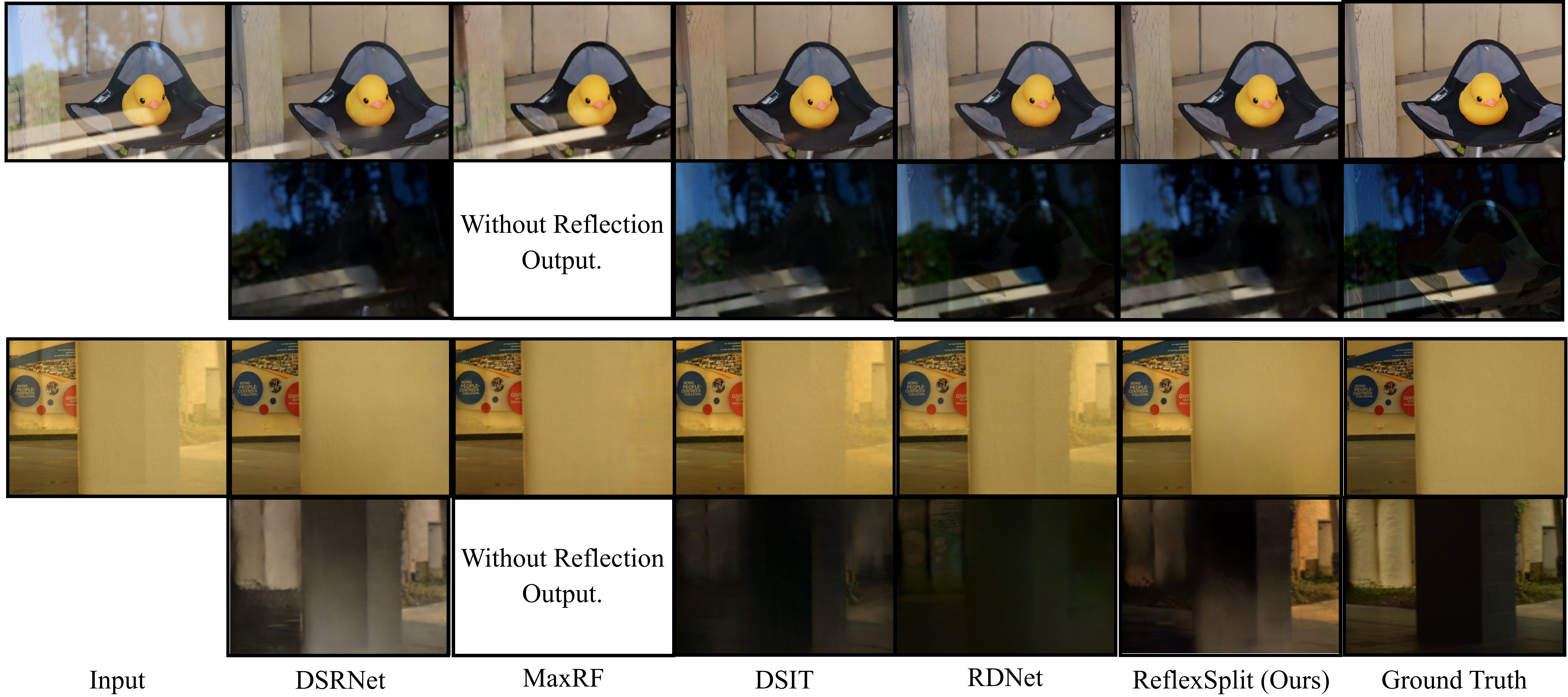}
    \caption{\textbf{Qualitative comparison on Wild and SolidObject~\cite{wan2017benchmarking}.}}
    \label{fig:wild_comparison}
    \vspace{-0.2cm}
\end{figure*}

\begin{figure*}[t]
\centering
\includegraphics[width=\textwidth]{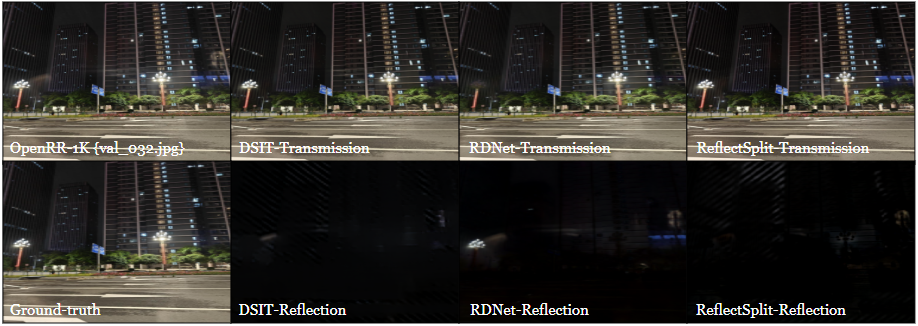}
\caption{\textbf{Additional visualization on OpenRR-1K~\cite{openrr1k}.}}
\label{fig:ReflexSplit_val032}
\end{figure*}

We provide additional visual comparisons on OpenRR-1K~\cite{openrr1k} and SIR$^2$~\cite{wan2017benchmarking} datasets in Figures~\ref{fig:openrr_additional}--\ref{fig:ReflexSplit_val032}.

As shown in Figure~\ref{fig:openrr_additional}, ReflexSplit achieves cleaner transmission layers in challenging real-world scenarios. Figures~\ref{fig:postcard_comparison} and~\ref{fig:wild_comparison} demonstrate consistent performance on synthetic benchmarks across different methods, each exhibiting different characteristics: DSRNet and DSIT show residual reflections in certain regions, MaxRF tends toward aggressive smoothing, while RDNet occasionally exhibits color shifts. Our method aims to balance these trade-offs through explicit layer separation constraints.

\section{Failure Cases and Limitations}
\label{sec:limitations}

\begin{figure*}[t]
\centering
\includegraphics[width=\textwidth]{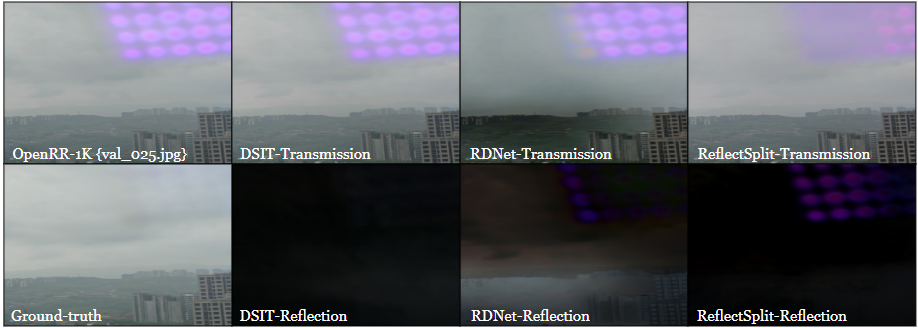}
\caption{\textbf{Failure cases on OpenRR-1K~\cite{openrr1k}.} 
}
\label{fig:ReflexSplit_val025_faliure}
\end{figure*}

\begin{figure*}[t]
\centering
\includegraphics[width=1\textwidth]{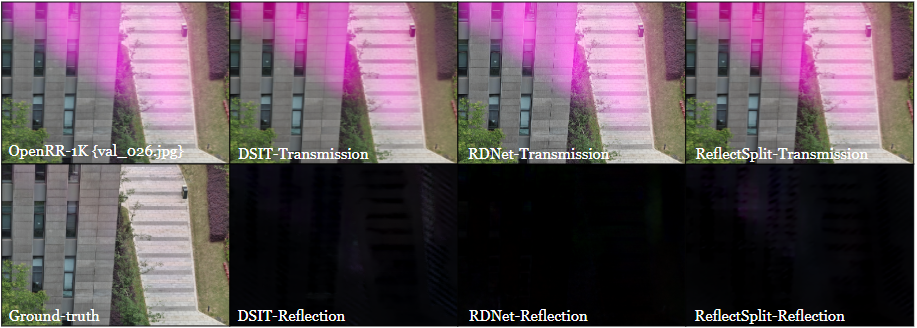}
\caption{\textbf{Failure cases on OpenRR-1K~\cite{openrr1k}.} }
\label{fig:failure_cases}
\end{figure*}

While ReflexSplit achieves state-of-the-art performance, we observe failure cases in challenging lighting conditions (Figures~\ref{fig:ReflexSplit_val025_faliure},~\ref{fig:failure_cases}):

\noindent\textbf{Complex outdoor lighting.} 
In outdoor scenes with strong sunlight, shadows, and varying illumination, the nonlinear mixing becomes highly spatially-varying. The separation becomes particularly challenging when transmission and reflection layers experience different lighting conditions simultaneously, causing inconsistent separation across image regions.

\noindent\textbf{Specular reflections.} 
Strong specular highlights from mirror-like surfaces produce localized over-exposure that saturates both layers. In these regions, the separation becomes ill-posed as no texture information remains to distinguish transmission from reflection—a fundamental challenge for learning-based approaches.

\noindent\textbf{Mixed indoor-outdoor scenes.} 
When capturing indoor scenes through glass with outdoor backgrounds, the extreme brightness difference (often exceeding camera dynamic range) causes either indoor under-exposure or outdoor over-saturation. Reliable layer separation becomes difficult when one component is dominated by noise or clipping artifacts, representing a limitation shared by most decomposition methods.

\section{Network Architecture Details}
\label{sec:architecture}
Table~\ref{tab:architecture} details ReflexSplit's complete architecture across four stages: dual-branch feature extraction (GFEB, LFEB), feature mixing, hierarchical decoding with CrGF and LFSB, and output generation.

\section{More Complexity Comparison}
Table~\ref{tab:efficiency} compares the computational efficiency of ReflexSplit 
against existing methods on $384 \times 384$ input resolution. ReflexSplit 
achieves a favorable balance between model complexity and task capability. 
While DAI and RDNet require multiple training stages and incur substantially 
higher parameter counts (1738.5M and 266.4M, respectively), ReflexSplit 
completes training in a single stage with 174M parameters, simplifying the 
training pipeline without sacrificing reflection separation capability. 
Compared to DExNet and DSIT, which are also single-stage methods supporting 
reflection separation, ReflexSplit achieves competitive FLOPs (969.050G vs. 
995.841G and 867.704G) while offering more comprehensive dual-branch feature 
modeling. Although MaxRF reports lower FLOPs (92.135G) and faster inference 
(24.48 FPS), it does not support direct reflection separation, limiting its 
applicability to this task. Overall, ReflexSplit represents a practical 
single-stage solution that targets reflection separation with reasonable 
computational overhead, making it suitable for deployment without the 
complexity of multi-stage training pipelines.

\begin{table}[t!]
\centering
\caption{Efficiency comparison on $384 \times 384$ input resolution.}
\vspace{-3mm}
\label{tab:efficiency}
\scalebox{0.65}{
\begin{tabular}{lcccccccc}
\toprule
\multirow{2}{*}{Method} 
& \multirow{2}{*}{Params (M)} 
& \multirow{2}{*}{FLOPs (G)} 
& \multirow{2}{*}{Time (ms)} 
& \multirow{2}{*}{FPS} 
& {Training}
& {Reflection} \\
& & & & & Stage& Separation\\
\midrule
DExNet & 9.6   & 995.841     & 185.114     & 5.40    & Single & \checkmark\\
MaxRF  & 27.9  & 92.135 & 40.855 & 24.48 & Multiple & $\times$ \\
DSIT   & 136   & 867.704     & 172.199     & 5.81    & Single & \checkmark\\
RDNet  & 266.4 & 1047.141     & 122.800     & 8.14    & Multiple & \checkmark\\
DAI    & 1738.5    & 7850.567   & 418.649     & 2.39    & Multiple & $\times$\\
\cellcolor{gray!20} \textbf{ReflexSplit}   & \cellcolor{gray!20} 174 & \cellcolor{gray!20} 969.050 & \cellcolor{gray!20} 211.349  & \cellcolor{gray!20} 4.73 & \cellcolor{gray!20} Single & \cellcolor{gray!20} \checkmark\\
\bottomrule
\end{tabular}
}
\end{table}

\begin{table*}[t]
\centering
\caption{\textbf{Architecture of ReflexSplit.} The model takes a $384\times384$ input image and processes it through dual-branch encoders (GFEB and LFEB) followed by hierarchical decoding with CrGF and LFSB.}
\label{tab:architecture}
\resizebox{0.77\textwidth}{!}{
\begin{tabular}{l|c|c|c|c}
\hline
\textbf{Block Name} & \textbf{Output Size} & \textbf{Transmission Branch} & \textbf{Reflection Branch} & \textbf{Order} \\
\hline
\hline
\multicolumn{5}{c}{\textit{Stage 1: Dual-Branch Feature Extraction}} \\
\hline
\textbf{Global Feature Extractor Block (GFEB)} & & & & \\
Swin Transformer Stage 2 & $96 \times 96 \times 192$ & \multicolumn{2}{c|}{Shared ($\mathbf{P}_2$)} & 1 \\
Swin Transformer Stage 3 & $48 \times 48 \times 384$ & \multicolumn{2}{c|}{Shared ($\mathbf{P}_3$)} & 2 \\
Swin Transformer Stage 4 & $24 \times 24 \times 768$ & \multicolumn{2}{c|}{Shared ($\mathbf{P}_4$)} & 3 \\
Swin Transformer Stage 5 & $12 \times 12 \times 1536$ & \multicolumn{2}{c|}{Shared ($\mathbf{P}_5$)} & 4 \\
\hline
\textbf{Local Feature Extractor Block (LFEB)} & & & & \\
Conv 3×3 & $384 \times 384 \times 48$ & $3 \rightarrow 48$ ($\mathbf{E}_0$) & $3 \rightarrow 48$ ($\mathbf{E}_0$) & 1 \\
MuGI Block (×2) & $384 \times 384 \times 48$ & $48 \rightarrow 48$ & $48 \rightarrow 48$ & 2 \\
Conv 2×2, stride=2 & $192 \times 192 \times 96$ & $48 \rightarrow 96$ ($\mathbf{E}_1$) & $48 \rightarrow 96$ ($\mathbf{E}_1$) & 3 \\
MuGI Block (×2) & $192 \times 192 \times 96$ & $96 \rightarrow 96$ & $96 \rightarrow 96$ & 4 \\
Conv 2×2, stride=2 & $96 \times 96 \times 192$ & $96 \rightarrow 192$ ($\mathbf{E}_2$) & $96 \rightarrow 192$ ($\mathbf{E}_2$) & 5 \\
MuGI Block (×2) & $96 \times 96 \times 192$ & $192 \rightarrow 192$ & $192 \rightarrow 192$ & 6 \\
Conv 2×2, stride=2 & $48 \times 48 \times 384$ & $192 \rightarrow 384$ ($\mathbf{E}_3$) & $192 \rightarrow 384$ ($\mathbf{E}_3$) & 7 \\
MuGI Block (×2) & $48 \times 48 \times 384$ & $384 \rightarrow 384$ & $384 \rightarrow 384$ & 8 \\
Conv 2×2, stride=2 & $24 \times 24 \times 768$ & $384 \rightarrow 768$ ($\mathbf{E}_4$) & $384 \rightarrow 768$ ($\mathbf{E}_4$) & 9 \\
MuGI Block (×2) & $24 \times 24 \times 768$ & $768 \rightarrow 768$ & $768 \rightarrow 768$ & 10 \\
Conv 2×2, stride=2 & $12 \times 12 \times 1536$ & $768 \rightarrow 1536$ ($\mathbf{E}_5$) & $768 \rightarrow 1536$ ($\mathbf{E}_5$) & 11 \\
\hline
\hline
\multicolumn{5}{c}{\textit{Stage 2: Feature Mixing}} \\
\hline
\textbf{Initial Feature Interaction} & & & & \\
PixelShuffle (×2) & $24 \times 24 \times 384$ & $1536 \rightarrow 384$ & $1536 \rightarrow 384$ & 5 \\
LFSB (depth=5) & $24 \times 24 \times 384$ & $384 \rightarrow 384$ & $384 \rightarrow 384$ & 6 \\
Conv 1×1 & $24 \times 24 \times 768$ & $384 \rightarrow 768$ & $384 \rightarrow 768$ & 7 \\
LFSB (depth=4) & $24 \times 24 \times 768$ & $768 \rightarrow 768$ & $768 \rightarrow 768$ & 8 \\
\hline
\hline
\multicolumn{5}{c}{\textit{Stage 3: Hierarchical Decoding with CrGF and LFSB}} \\
\hline
\textbf{Decoder Level 4} & & & & \\
CrGF & $24 \times 24 \times 768$ & $\mathbf{P}_4 + \mathbf{E}_4 + \mathbf{F}_{5}$ & $\mathbf{P}_4 + \mathbf{E}_4 + \mathbf{F}_{5}$ & 9 \\
LFSB (×12, depth=4) & $24 \times 24 \times 768$ & $768 \rightarrow 768$ & $768 \rightarrow 768$ & 10 \\
PixelShuffle (×2) & $48 \times 48 \times 192$ & $768 \rightarrow 192$ & $768 \rightarrow 192$ & 11 \\
MuGI Block (×2) & $48 \times 48 \times 192$ & $192 \rightarrow 192$ & $192 \rightarrow 192$ & 12 \\
Conv 1×1 & $48 \times 48 \times 384$ & $192 \rightarrow 384$ & $192 \rightarrow 384$ & 13 \\
\hline
\textbf{Decoder Level 3} & & & & \\
CrGF & $48 \times 48 \times 384$ & $\mathbf{P}_3 + \mathbf{E}_3 + \mathbf{F}_4$ & $\mathbf{P}_3 + \mathbf{E}_3 + \mathbf{F}_4$ & 14 \\
LFSB (depth=3) & $48 \times 48 \times 384$ & $384 \rightarrow 384$ & $384 \rightarrow 384$ & 15 \\
LFSB (×8, depth=3) & $48 \times 48 \times 384$ & $384 \rightarrow 384$ & $384 \rightarrow 384$ & 16 \\
PixelShuffle (×2) & $96 \times 96 \times 96$ & $384 \rightarrow 96$ & $384 \rightarrow 96$ & 17 \\
MuGI Block (×2) & $96 \times 96 \times 96$ & $96 \rightarrow 96$ & $96 \rightarrow 96$ & 18 \\
Conv 1×1 & $96 \times 96 \times 192$ & $96 \rightarrow 192$ & $96 \rightarrow 192$ & 19 \\
\hline
\textbf{Decoder Level 2} & & & & \\
CrGF & $96 \times 96 \times 192$ & $\mathbf{P}_2 + \mathbf{E}_2 + \mathbf{F}_3$ & $\mathbf{P}_2 + \mathbf{E}_2 + \mathbf{F}_3$ & 20 \\
LFSB (depth=2) & $96 \times 96 \times 192$ & $192 \rightarrow 192$ & $192 \rightarrow 192$ & 21 \\
LFSB (×4, depth=2) & $96 \times 96 \times 192$ & $192 \rightarrow 192$ & $192 \rightarrow 192$ & 22 \\
PixelShuffle (×2) & $192 \times 192 \times 48$ & $192 \rightarrow 48$ & $192 \rightarrow 48$ & 23 \\
MuGI Block (×2) & $192 \times 192 \times 48$ & $48 \rightarrow 48$ & $48 \rightarrow 48$ & 24 \\
Conv 1×1 & $192 \times 192 \times 96$ & $48 \rightarrow 96$ & $48 \rightarrow 96$ & 25 \\
\hline
\textbf{Decoder Level 1} & & & & \\
Direct Aggregation & $192 \times 192 \times 96$ & $\mathbf{E}_1 + \mathbf{F}_2$ & $\mathbf{E}_1 + \mathbf{F}_2$ & 26 \\
LFSB (×2, depth=1) & $192 \times 192 \times 96$ & $96 \rightarrow 96$ & $96 \rightarrow 96$ & 27 \\
PixelShuffle (×2) & $384 \times 384 \times 24$ & $96 \rightarrow 24$ & $96 \rightarrow 24$ & 28 \\
MuGI Block (×2) & $384 \times 384 \times 24$ & $24 \rightarrow 24$ & $24 \rightarrow 24$ & 29 \\
Conv 1×1 & $384 \times 384 \times 48$ & $24 \rightarrow 48$ & $24 \rightarrow 48$ & 30 \\
\hline
\textbf{Decoder Level 0} & & & & \\
Direct Aggregation & $384 \times 384 \times 48$ & $\mathbf{E}_0 + \mathbf{F}_1$ & $\mathbf{E}_0 + \mathbf{F}_1$ & 31 \\
LFSB (×2, depth=0) & $384 \times 384 \times 48$ & $48 \rightarrow 48$ & $48 \rightarrow 48$ & 32 \\
MuGI Block (×2) & $384 \times 384 \times 48$ & $48 \rightarrow 48$ & $48 \rightarrow 48$ & 33 \\
\hline
\hline
\multicolumn{5}{c}{\textit{Stage 4: Output Generation}} \\
\hline
Conv 3×3 & $384 \times 384 \times 3$ & $48 \rightarrow 3$ ($\hat{\mathbf{T}}$) & $48 \rightarrow 3$ ($\hat{\mathbf{R}}$) & 34 \\
\hline
\textbf{Learnable Residue Module (LRM)} & & & & \\
Feature Aggregation & $384 \times 384 \times 48$ & \multicolumn{2}{c|}{$\mathbf{F}^t_0 + \mathbf{F}^r_0$} & 35 \\
SinBlock & $384 \times 384 \times 48$ & \multicolumn{2}{c|}{$48 \rightarrow 48$} & 36 \\
Conv 3×3 + Tanh & $384 \times 384 \times 3$ & \multicolumn{2}{c|}{$48 \rightarrow 3$ ($\hat{\mathbf{RR}}$)} & 37 \\
\hline
\end{tabular}
}
\end{table*}

{
    \small
    \bibliographystyle{ieeenat_fullname}
    \bibliography{main}
}

\end{document}